\PassOptionsToPackage{table}{xcolor}
\documentclass[10pt]{article}
\usepackage{natbib}
\usepackage[utf8]{inputenc}
\usepackage[T1]{fontenc}
\usepackage{amsmath}
\usepackage{multicol}
\usepackage{algorithm}
\usepackage{algpseudocode}
\usepackage{subfigure}
\usepackage{caption}
\usepackage{subcaption}
\usepackage{array}
\usepackage[table]{xcolor}
\usepackage{multirow}
\usepackage{hyperref}
\usepackage{graphicx}
\usepackage[format=plain,justification=justified]{caption}
\usepackage{float}
\usepackage{booktabs}
\usepackage{authblk} 
\allowdisplaybreaks

\title{
A Multi-Centric Anthropomorphic 3D CT Phantom-Based Benchmark Dataset for Harmonization
}

\newcolumntype{C}[1]{>{\centering\arraybackslash}p{#1}}

\author[1,2]{Mohammadreza Amirian}
\author[3]{Michael Bach}
\author[6]{Oscar Jimenez-del-Toro}
\author[3]{Christoph Aberle}
\author[1]{Roger Schaer}
\author[1,2]{Vincent Andrearczyk}
\author[1]{Jean-Félix Maestrati}
\author[1]{Maria Martin Asiain}
\author[5]{Kyriakos Flouris}
\author[3]{Markus Obmann}
\author[7]{Clarisse Dromain}
\author[8]{Benoît Dufour}
\author[4]{Pierre-Alexandre Alois Poletti}
\author[9]{Hendrik von Tengg-Kobligk}
\author[10]{Rolf Hügli}
\author[11]{Martin Kretzschmar}
\author[12]{Hatem Alkadhi}
\author[5]{Ender Konukoglu}
\author[1,4]{Henning Müller}
\author[3]{Bram Stieltjes}
\author[1,2,*]{Adrien Depeursinge}

\affil[1]{Institute of Informatics, School of Management, HES-SO Valais-Wallis, Sierre, Switzerland}
\affil[2]{Nuclear Medicine and Molecular Imaging Department, Lausanne University Hospital, Lausanne, Switzerland}
\affil[3]{Clinic of Radiology and Nuclear Medicine, University Hospital Basel, Basel, Switzerland}
\affil[4]{Faculty of Medicine, University of Geneva (UNIGE), Geneva, Switzerland}
\affil[5]{Computer Vision Lab, ETH Zurich, Zurich, Switzerland}
\affil[6]{Idiap Research Institute, Martigny, Switzerland}
\affil[7]{Department of Radiology, Lausanne University Hospital, Lausanne, Switzerland}
\affil[8]{Groupe 3R, Lausanne-Épalinges Imaging Center, Lausanne, Switzerland}
\affil[9]{Inselspital Bern, University of Bern, Bern, Switzerland}
\affil[10]{Cantonal Hospital Baselland, Bruderholz, Switzerland}
\affil[11]{Schmerzklinik Basel, Basel, Switzerland}
\affil[12]{Diagnostic and Interventional Radiology, University Hospital Zurich, Switzerland}
\affil[*]{Corresponding author: adrien.depeursinge@hevs.ch}

\date{}

\begin{document}

\flushbottom
\maketitle
\newpage
\begin{abstract}
Artificial intelligence (AI) has introduced numerous opportunities for human assistance and task automation in medicine. However, it suffers from poor generalization in the presence of shifts in the data distribution. In the context of AI-based computed tomography (CT) analysis, significant data distribution shifts can be caused by changes in scanner manufacturer, reconstruction technique or dose.
AI harmonization techniques can address this problem by reducing distribution shifts caused by various acquisition settings. This paper presents an open-source benchmark dataset containing CT scans of an anthropomorphic phantom acquired with various scanners and settings, which purpose is to foster the development of AI harmonization techniques. Using a phantom allows fixing variations attributed to inter- and intra-patient variations.
The dataset includes 1378 image series acquired with 13 scanners from 4 manufacturers across 8 institutions using a harmonized protocol as well as several acquisition doses.
Additionally, we present a methodology,  baseline results and open-source code to assess image- and feature-level stability and liver tissue classification, promoting the development of AI harmonization strategies.

\end{abstract}

\thispagestyle{empty}


\section*{Background \& Summary}

Recent breakthroughs in data-driven algorithms and artificial intelligence (AI) applications in medical information processing have introduced tremendous potential for AI-assisted image-based personalized medicine that addresses tasks such as segmentation, diagnosis, and prognosis~\cite{lambin2017radiomics}. 
However, these opportunities come with two challenges: large data requirements and consistency in data distribution.
Machine and deep learning algorithms have extreme data demand, which is coupled with the high costs of data acquisition, and annotation for a single observation (e.g., one event corresponds to one patient in a survival study).
These challenges encourage pooling of data collected from multiple centers and scanners to achieve a critical mass of data for training models. 
However, pooling data from multiple centers introduces significant variability in the acquisition parameters and specifics of image reconstruction algorithms, leading to data domain shifts and inconsistencies in the collected data.
The domain shift introduced by this variability in scanners reduces the value of merging data from multiple centers, reducing performance of predictive tasks such as segmentation, diagnosis, and prognosis, as well as in federated scenarios.
Furthermore, domain shifts between training and test or inference data entails high risks of incorrect and uncontrolled predictions for treatment planning and personalized medicine when the inference is based on a scanner (and/or acquisition setting) that was not represented in the training data.
Although this challenge applies to all medical imaging modalities, it is particularly important for computed tomography (CT) images due to the wide range of variability in manufacturers, acquisition parameters and dose, reconstruction algorithms, and customized parameter tunings in different centers.

Since the beginning of the development of recent revolutionary data-driven modeling using machine and deep learning, generalization to unseen data to guarantee practical usage of the developed models has been a serious issue~\cite{neyshabur2017exploring}.
Model generalization was an even more serious issue between domains in the presence of changes in data distribution caused by diverse data collection settings~\cite{hendrycks2021many}.
Furthermore, deploying the models developed in practical applications demonstrated that generalization to training distribution data is not sufficient because the data distribution can also change over time~\cite{guo2022evaluation}.
Poor generalization in the presence of data distribution shifts has encouraged considerable research endeavors in transfer learning, lifelong learning, and (one- or few-shot) domain adaptation to improve the models' generalization and increase their robustness against changes in data.
The body of research on adapting models to cope with data variations is extensive; however, collecting and developing dedicated real-world datasets to study domain shift in a controlled fashion and processing the data to minimize the discrepancies has received much less attention in the literature.

\begin{figure}[ht!]
     \centering
     \includegraphics[width=0.7\textwidth]{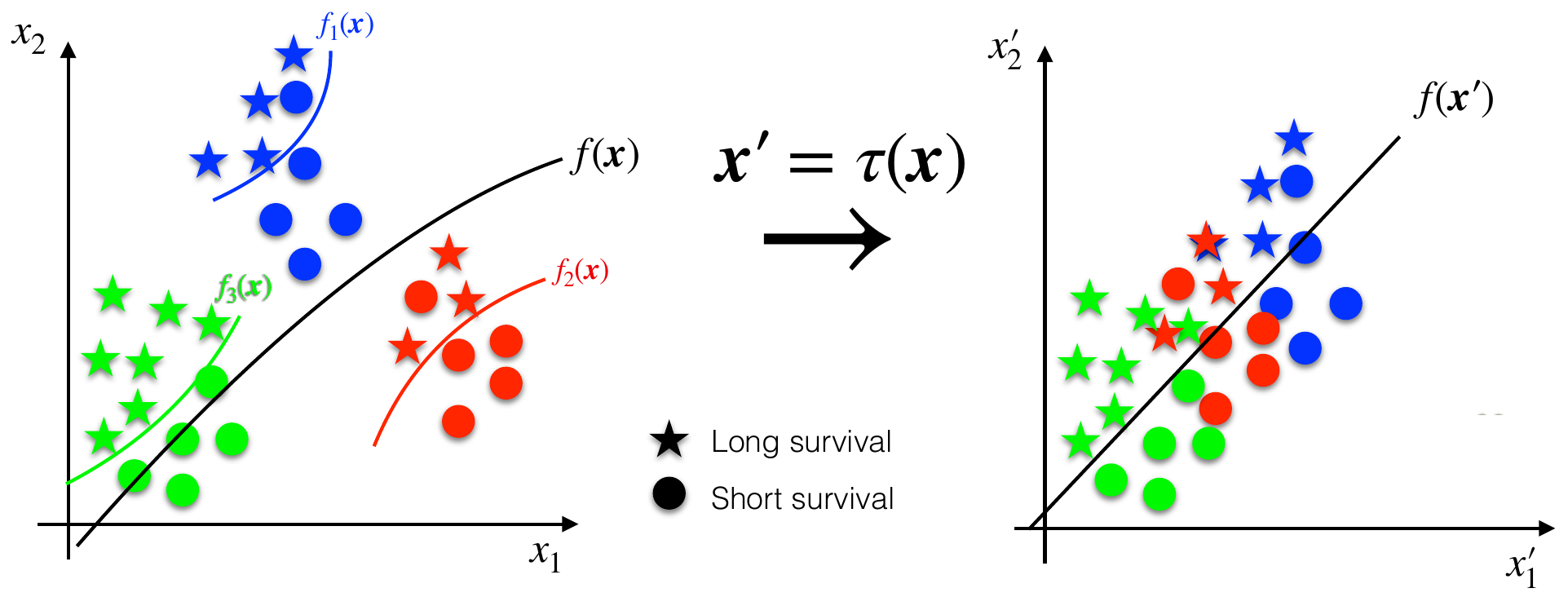}

    \caption{Visual representation of the goal of harmonization techniques as defined and stated in this research work. The data points from different data sources are presented in different colors for the long versus short survival classification task, represented by stars and circles. (Left) The functions $f_i(\boldsymbol{x})$ represent decision rule for the suitable classifier for the survival prediction task for each dataset $i$ (e.g. acquired with scanner $i$) based on the feature vector $\boldsymbol{x}$. (Right) The harmonization transformation $\tau$ aims to map the original feature vector $\boldsymbol{x}$ to a harmonized one $\boldsymbol{x'}=\tau(\boldsymbol{x})$ such that one single global classifier $f(\boldsymbol{x})$ can be used for the survival task across all data sources.}
    \label{fig:harmonization}
\end{figure}

Research works have addressed the challenge of data distribution shifts in raw data, and feature domains.
Data distribution shifts can be targeted by preprocessing the raw data before they are presented to the models instead of updating and adapting the models.
The literature dealing with reducing discrepancies from input data is presented, although not exclusively, under the topic of data harmonization.
Researchers also aimed to develop models that are robust to domain shifts in image and feature space (image- and feature-level) by adapting the model based on shifts in the feature space.
In this paper, we refer to harmonization as data alignment techniques~\cite{mali2021making}, either in image or feature domains, such that the models' performance does not change from one data distribution to another for a given task such as tissue classification or segmentation as visually depicted in Fig.~\ref{fig:harmonization}.
We consider a machine learning function $f_i(\boldsymbol{x})$ optimized on a dataset $i$ (e.g. acquired using scanner $i$) to perform a task such as classification (e.g. diagnosis), survival (e.g. prognosis), or segmentation based on a feature vector $\boldsymbol{x}$ extracted from a CT image $I$.
In addition, we consider a harmonization transformation $\tau(\boldsymbol{x})$ that aims to map the original feature vector $\boldsymbol{x}$ to a harmonized one  $\boldsymbol{x}'=\tau(\boldsymbol{x})$ such that a single global classifier $f(\boldsymbol{x})$ can be used for the task across heterogeneous pooled data sources.
Image-level harmonization techniques (also referred to as ``homogenization") refer to the methods aiming at removing discrepancies between data from several sources~\cite{amirian2021prepnet} in the original image space.
The goal of harmonization algorithms is finding a transformation $\tau(I)$ to transform the original CT scan $I$ to a harmonized one $I'=\tau(I)$ such that one single modeled function $f(I)$ can be efficiently learned from a pooled collection of images acquired from different sources. 

There is a wide range of methods for harmonization both at image- and feature-level in the literature.
ComBat presents an analytical solution to feature-level data harmonization based on a theory introduced first in the field of genomics~\cite{orlhac2022guide} to remove sequencing batch effects.
In the context of deep learning (DL) research, gradient-based methods such as presented in~\cite{andrearczyk2019learning, andrearczyk2019neural, amirian2021prepnet} as well as generative adversarial networks (GANs~\cite{tor2020unsupervised,liu2021style,sinha2021alzheimer}) and cycle-consistent GANs~\cite{Bashyam2020MedicalIH,marcadent2020generative} are used for mapping and unifying data across several acquisition setups.
There are numerous studies in the literature investigating feature stability in radiomics~\cite{mali2021making,hertel2023phantom,mackin2015measuring}, which highlight the strong consequences of image acquisition on feature values.
The main motivation behind measuring feature stability is related to the performance of a given clinical task (e.g. diagnosis, prognosis or segmentation).
However, this measure is irrelevant when taken alone as long as the performed clinical task is not significantly affected by these variations.
For instance, for a diagnostic task defined as classification based on a single feature, if intra-class variation due to different scanners is smaller than the inter-class variations, the results remain unchanged.
The impact of stability on an actual clinical task was rarely investigated with only a handful of studies available~\cite{Michallek2022,Zhang2023,jimenez2021discriminative}.

Despite valuable contributions in data harmonization subjects~\cite{mali2021making}, focused open-source benchmark datasets with predefined evaluation to compare the merits of these techniques are scarce in the literature.
In particular, datasets that can allow to disentangle variations coming from disease, anatomical or physiological changes from the variations attributed to imaging acquisition hardware or reconstruction parameters are needed. 
Test-retest approaches specifically address this need~\cite{AVL2014,Fedorov2018}, and they consist of repeatedly scanning the same object of interest while carefully controlling and modifying data acquisition parameters. 
For imaging, phantoms offer ideal objects of interest for the test-retest methods allowing large numbers of repetitions and avoiding patient irradiation. 
Ideally, the phantoms should closely mimic human tissue and organs containing potential regions of interest (ROI) for relevant diseases.
A range of materials including cartridges~\cite{mackin2015measuring, aymerich2023pilot}, and bio-organic substances such as fruits~\cite{hertel2023phantom} and meat~\cite{wang2023radiomic} have been used to form phantoms to scan instead of patients' bodies.
However, 3D prints of real patient anatomy become more popular in recent research works~\cite{jahnke2017radiopaque} thanks to their anthropomorphic nature.

This paper presents a dataset of CT scans from a 3D-printed iodine-ink anthropomorphic phantom to encourage research in AI harmonization.
Future studies can develop harmonization methods and report their performance using this dataset for unbiased comparison; consequently, this dataset can serve as a benchmark to advance scientific research in harmonization.
We present a dataset of $268$ CT image series acquired with $13$ different scanners by $4$ different manufacturers at $8$ institutions with multiple imaging setups (see Table~\ref{table:scaners}).
These scans are acquired using a harmonized protocol (explained in more detail in the next section) defined based on surveying actual protocols used in institutions and using a dose ($\text{CTDI}_{\text{vol}}$) of $10~mGy$. 
In addition to this main collection, we repeated the acquisition with various doses including $1~mGy$, $3~mGy$, $6~mGy$, and $14~mGy$, resulting in a dataset of 1378 CT images series (see Table~\ref{table:scans}) from 649 CT scans.
The scans are conducted using a phantom based on real human CT acquisitions~\cite{bach20233d, jahnke2017radiopaque}, and the impact of different manufacturers and scanners is shown in the image features and tissue classification task.
The scanned phantom not only contains a 3D structure printed with iodine ink on paper with realistic human tissue texture focusing on the liver but also includes a thoracic region and synthetic test patterns~\cite{bach20233d}.
For each scan, the liver region includes six ROIs with masks from four different tissue classes.
This setup enables the evaluation of harmonization techniques not only for their effect on feature stability but also in the context of a tissue classification task.
The dataset is publicly available as part of the cancer imaging archive (TCIA) collection\footnote{\url{https://www.cancerimagingarchive.net/collection/ct4harmonization-multicentric/}}.

The remainder of this paper presents the details of the harmonized acquisition protocol, a description of centers where the phantom is scanned, and explanations of the phantom creation. This is followed by presenting sample images and introducing evaluation metrics, demonstration of the dataset's applicability, relevance, and baseline results.
We discuss the details of the data organization, followed by a brief description of its usage with an open-source code repository. 
This research work also presents a standard evaluation method with defined metrics and data splits, followed by baseline results without any harmonization, which future research can aim to improve.
Finally, this paper concludes with a discussion of the limitations of the presented dataset and potential improvements for future work.



\section*{Methods}\label{sec:methods}
This section presents the methodology employed for collecting the dataset, along with the methods and metrics proposed to evaluate the performance of the harmonization techniques that will be proposed in future research based on this work.
The dataset presented in this study focuses primarily on the impact of CT scanners and their manufacturers, both on acquired images and derived quantitative features. 
To this end, a harmonized imaging protocol with a fixed radiation dose\footnote{Additional doses levels are also available.}, acquisition and reconstruction parameters was used.
Anatomical, physiological, and disease-related variations are not present thanks to the use of a fixed anthropomorphic CT phantom.

\subsection*{Anthropomorphic Phantom}
As a surrogate for the human body, a 3D-printed iodine-ink, paper-based CT phantom is used in this study.
To create the phantom as realistically and closely as possible to human anatomy, it was manufactured based on a human body CT scan, focusing on the liver region and including six ROIs from four distinct tissue classes.
Iodine ink was injected into the paper to increase its density to match the real tissue density in the reference human body~\cite{jahnke2017radiopaque}.
The phantom is detailed and evaluated in Bach et al.~\cite{bach20233d}.

\begin{figure}[ht!]
    \centering
    \begin{tabular}{cc}
         \includegraphics[width=0.5\textwidth]{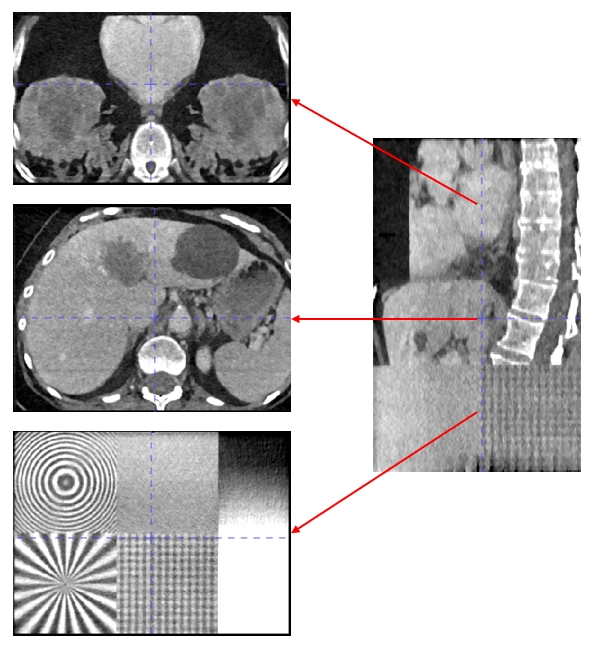}
    \end{tabular}    
    \caption{Axial and sagittal views of the three segments of the anthropomorphic phantom including pulmonary, liver, and test patterns (level=50, window=400).}
    \label{fig:phantom}
\end{figure}

\begin{figure}[ht!]
    \centering  
    \subfigure[Phantom]{
    \includegraphics[width=0.45\textwidth]{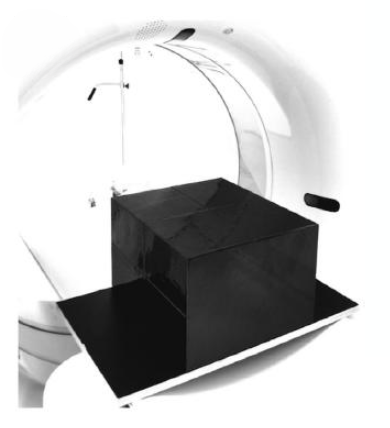}
    \label{fig:samplectphantom}} 
    \subfigure[3D rendering of a sample CT scan]{
    \includegraphics[width=0.45\textwidth]{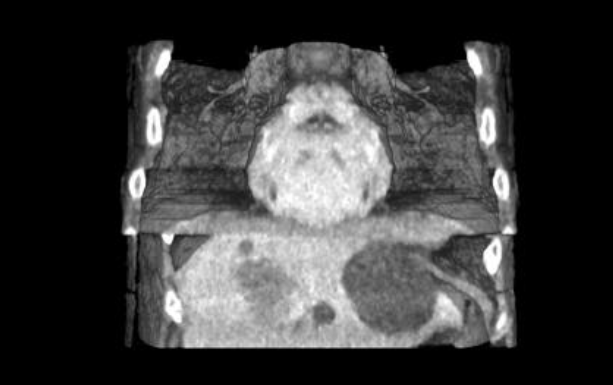}
    \label{fig:samplectscan}} \\
    \subfigure[Axial view of ROIs in liver tissue]{
    \includegraphics[width=0.45\textwidth]{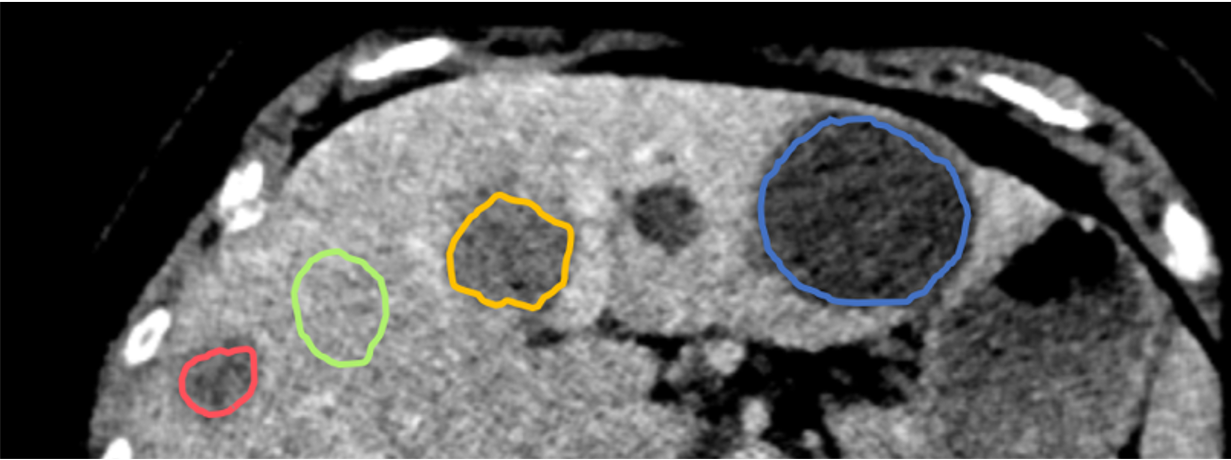}
    \label{fig:samplectregion1}}
    \subfigure[3D view of ROIs in liver tissue]{
    \includegraphics[width=0.45\textwidth]{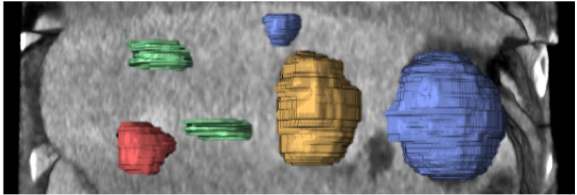}
    \label{fig:samplectregion2}} \\
    
    \subfigure[First cyst]{
    \includegraphics[width=0.15\textwidth]{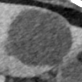}
    \label{fig:samplectcyst1}}
    \subfigure[Second cyst]{
    \includegraphics[width=0.15\textwidth]{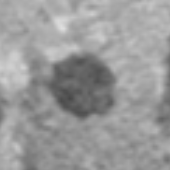}
    \label{fig:samplectcyst2}}
    \subfigure[Hemangioma]{
    \includegraphics[width=0.15\textwidth]{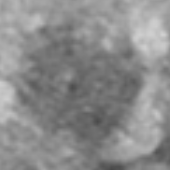}
    \label{fig:samplectHemangioma}}
    \subfigure[Metastasis]{
    \includegraphics[width=0.15\textwidth]{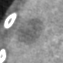}
    \label{fig:samplectmetastatsis}}
    \subfigure[First normal region]{
    \includegraphics[width=0.15\textwidth]{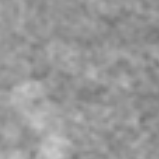}
    \label{fig:samplectnormal1}}
    \subfigure[Second normal region]{
    \includegraphics[width=0.15\textwidth]{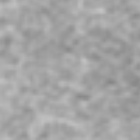}
    \label{fig:samplectnormal2}}
    \caption{Visual overview of the anthropomorphic CT phantom used in this study. a) Phantom and scanner (adapted from Bach et al.~\cite{bach20233d}). b) Sample CT scan. c)-d) Example of segmentation (6 ROIs) of liver tissue proposed by human experts with 4 classes (from Jimenez-del-Toro et al.~\cite{jimenez2021discriminative}): cyst (blue), hemangioma (yellow), metastasis from a colon carcinoma (red) and normal (green) . e)-j) Axial views of the six ROIs acquired using the SOMATOM Definition Edge scanner from Siemens with a slice thickness of $2~mm$ (level=50, window=400).}
    \label{fig:samplect}
\end{figure}

The advantage of using this phantom is that patient anatomy, physiology and disease could be fixed over the period of the experiments; hence any diversity in reconstructed images, and computed results only reflect the impact of the scanners or acquisition protocol on the images.
However, the minimum attenuation of the phantom is limited to the attenuation of paper, which is a limitation for organs containing low-density tissues or substances such as air in the lung. 
Therefore, this study focuses on the liver region.
The phantom includes three segments as depicted in Fig.~\ref{fig:phantom}: (i) thoracic, (ii) liver, and (iii) test patterns.
The liver section includes six ROIs from four classes including two cysts, a metastasis, a hemangioma, and two normal liver tissue regions which can be used for the classification of four classes, and for feature stability analysis (see Fig.~\ref{fig:samplect}).

\subsection*{Harmonized Acquisition Protocol}
Before the CT scans of the phantom were acquired, a survey was carried out to collect realistic acquisition and reconstruction parameter settings that are used in clinical thoracoabdominal CT scans for oncological staging, tumor search, and infectious foci in the portal venous contrast phase.
The survey includes 21 CT scanners from 9 centers across Switzerland. 
From the survey, a harmonized protocol based on averaged parameters was derived, i.e. a set of acquisition and reconstruction parameters representing typical, realistic clinical settings that are possible to be set on most scanners.
\label{sec:harmonizedprotocol}
\subsection*{Scanners and Centers}
After the harmonized protocol was defined, the anthropomorphic phantom was scanned on 13 different CT scanners at 8 Swiss centers (A-H, see Table~\ref{table:scaners}), including all five university hospitals, one cantonal hospital, and two private clinics. 
Newer and older models from four CT manufacturers were included to cover a wide variety of CT scanners.

Due to vendor-specific limitations, it was not possible to set exactly the same parameters on all CT scanners, so the parameters differ slightly from the harmonized protocol in some cases.
The actual acquisition and reconstruction parameter settings of all 13 CT scanners are listed in Table~\ref{table:scaners} for one of the dose levels (CTDI$_{\text{vol}}$ = 10 mGy).
CT scans were performed at five dose levels (1 mGy, 3 mGy, 6 mGy, 10 mGy, 14 mGy). 
Only the tube current time product (in mAs) was adjusted to set the various dose levels, all other parameters were kept the same.
For each CT scanner and for each dose level, 10 repeated scans with identical settings were performed, except inadvertently for the Toshiba Aquilion Prime SP scanner at 10 mGy (9 repetition scans). Thus, 649 CT scans were performed in total.

Images were reconstructed using two or three different reconstruction algorithms per CT scan, resulting in two or three CT image series per CT scan.
For all CT scans, a vendor-specific iterative reconstruction (IR) algorithm with a standard soft tissue kernel was used, resulting in 649 IR CT series.
In addition, filtered backprojection (FBP) reconstruction with a standard soft tissue kernel was used for all CT scans, resulting in another 649 FBP CT series.
For 2 of the 13 CT scanners, a DL based reconstruction algorithm was available.
For one of these scanners, it was used for three dose levels (1 mGy, 3 mGy, 6 mGy), resulting in 30 additional CT series.
For the second scanner, DL reconstruction was used for all five dose levels, resulting in 50 additional CT series.

In summary, the dataset presented in this work consists of 1378 series reconstructed from 649 CT scans. Table~\ref{table:scans} lists the number of available image series for each CT scanner.

%

\subsection*{Measuring Data Shifts and their Impact on Tissue Classification}
 
\begin{figure}[ht!]
    \centering
    \resizebox{\textwidth}{!}{
    \begin{tabular}{>{\centering\arraybackslash}m{3cm}|>{\centering\arraybackslash}m{3cm}|>{\centering\arraybackslash}m{3cm}|>{\centering\arraybackslash}m{3cm}|>{\centering\arraybackslash}m{3cm}}
  \toprule
  Siemens SOMATOM Definition Edge & Siemens SOMATOM Definition Flash & Siemens SOMATOM X.Cite & Siemens SOMATOM Edge Plus & Siemens SOMATOM Definition Edge \\
  \midrule
  \includegraphics[width=3cm]{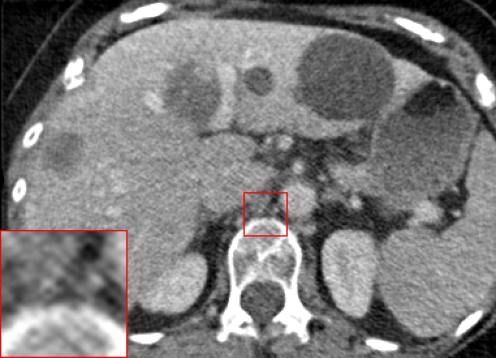}
  & \includegraphics[width=3cm]{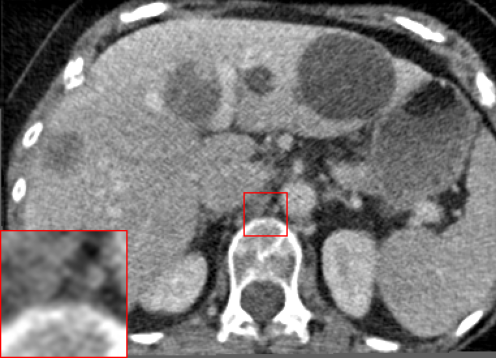}
  & \includegraphics[width=3cm]{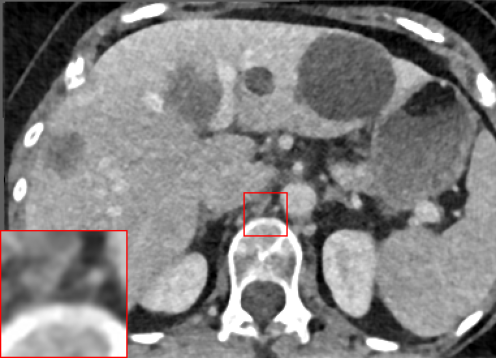}
  & \includegraphics[width=3cm]{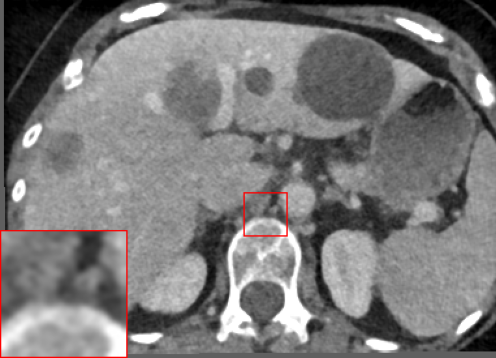}
  & \includegraphics[width=3cm]{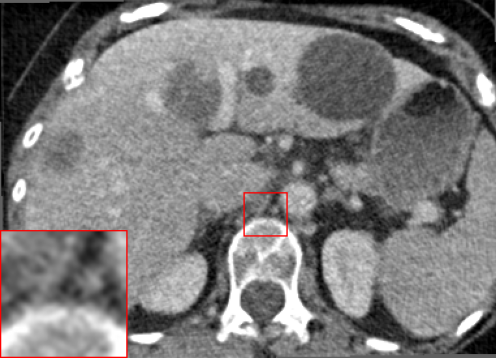} \\
  \midrule
  Siemens SOMATOM Definition Flash & Philips Brilliance iCT & Philips Brilliance CT 64 & GE Revolution Evo & GE Revolution Apex \\
  \midrule
  \includegraphics[width=3cm]{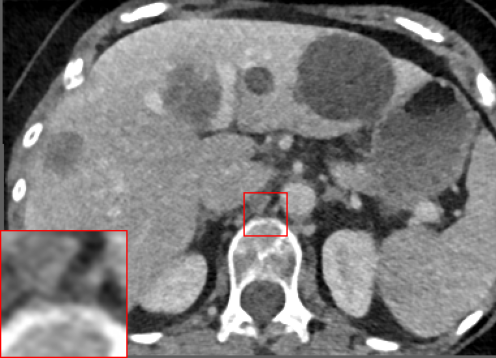}
  & \includegraphics[width=3cm]{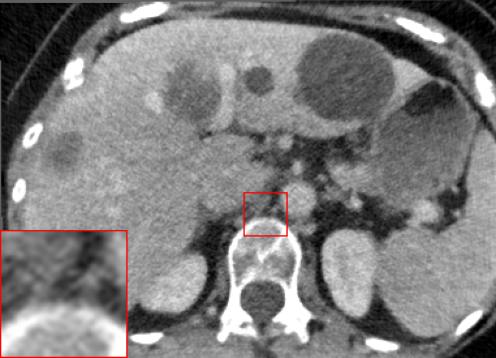}
  & \includegraphics[width=3cm]{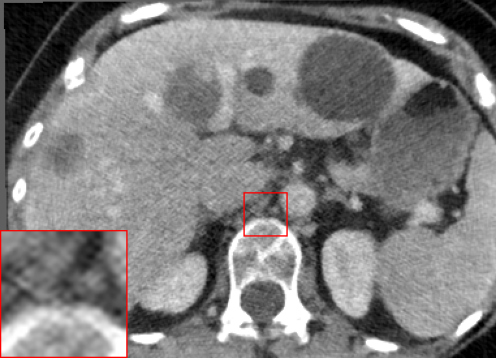}
  & \includegraphics[width=3cm]{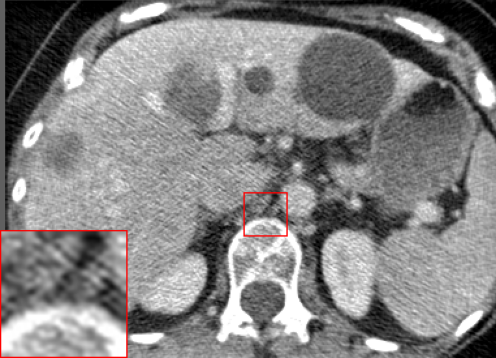}
  & \includegraphics[width=3cm]{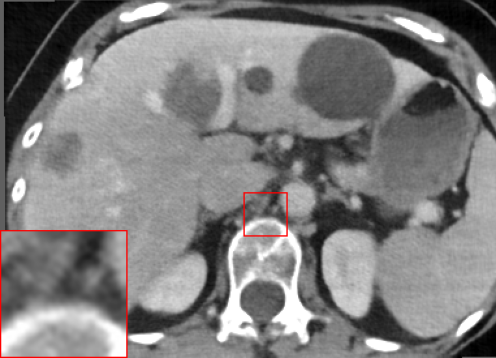} \\
  \midrule
  & GE BrightSpeed & Toshiba Aquilion Prime SP & Toshiba Aquilion CXL & \\
  \midrule
  
  & \includegraphics[width=3cm]{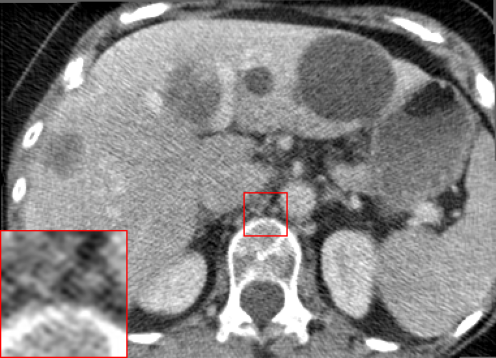}
  & \includegraphics[width=3cm]{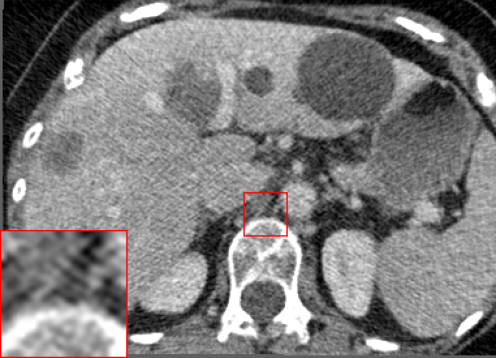}
  & \includegraphics[width=3cm]{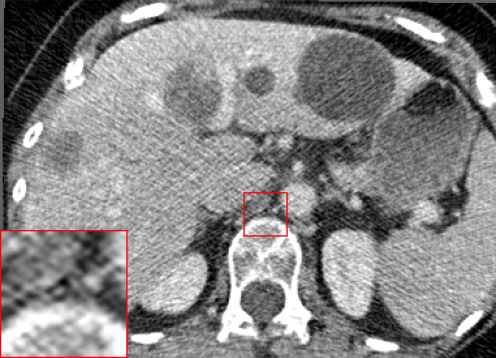}
  & \\
  \bottomrule
\end{tabular}

}
    \caption{Visual comparison of the texture in registered CT series reconstructed using iterative reconstruction (IR) from various scanners acquired using the harmonized protocol with a dose of 10 mGy (level=50, window=400).}
    \label{fig:scanner_comparison}
\end{figure}

Besides the technical information on the presented dataset, we suggest measures to quantify the stability of images and quantitative features to assess the performance of data harmonization methods.
Visually, the impact of scanner manufacturers is already evident in the texture of the acquired scans as depicted in Fig.~\ref{fig:scanner_comparison}, where scans are aligned using a rigid registration method to focus solely on the texture differences and to ignore minor positional shifts.
For image level stability assessment, the difference between a reference image series $I_r$ and a registered image series $I_s$ acquired with a scanner $s$ in the 3D volume domain can be measured using root mean square error (RMSE), peak signal-to-noise ratio (PSNR), or structural similarity~\cite{wang2004image} (SSIM). 
RMSE presents a simple measure focusing on voxel values, PSNR focuses on image quality by approximating the noise level, and SSIM is a visual quality measure reflecting structural similarities. 
These complementary metrics simultaneously estimate the pixel-level consistency, the level of noise introduced by changing scanner and acquisition settings, as well as structural consistencies, and are defined as
\begin{align}
    \text{RMSE}(I_r, I_s) &= \sqrt{\frac{1}{K} \sum_{k=1}^{K} \left( I_r[k] - I_s[k] \right)^2},\label{eq:RMSE} \\
    \text{PSNR}(I_r, I_s) &= 10 \cdot \log_{10} \left(\frac{{\max^2(I_r, I_s)}}{\text{MSE}(I_r, I_s)}\right),\label{eq:PSNR} \\
    \text{SSIM}(I_r, I_s) &= \frac{(2\mu_r\mu_s + C_1)(2\text{cov}_{rs} + C_2)}{(\mu_r^2 + \mu_s^2 + C_1)(\sigma_r^2 + \sigma_s^2 + C_2)}, \label{eq:SSIM}
\end{align}
where $K$ is the number of voxels, $\max(I_r, I_s)$ is the maximum voxel value among the two image series, MSE is the square of RMSE, $\mu_r$ and $\mu_s$ represent the mean intensities of image series $I_r$ and $I_s$, respectively, and their variances are denoted by $\sigma_r^2$ and $\sigma_s^2$.
The covariance between images is defined by $\text{cov}_{rs}$.
Given the measured dynamic range of CT scanners ($L = 3000~\text{HU}$), spanning from air ($-1000~\text{HU}$) to bone ($+2000~\text{HU}$), and the standard parameters of $K_1 = 0.01$ and $K_2 = 0.03$ proposed by Wang et al.~\cite{wang2004image}, we compute $C_1 = (K_1 \times L)^2 = 900~\text{HU}^2$ and $C_2 = (K_2 \times L)^2 = 8100~\text{HU}^2$.
It is worth noting that computing~\eqref{eq:RMSE},~\eqref{eq:PSNR} and~\eqref{eq:SSIM} requires to register $I_r$ and $I_s$, which itself has an impact on image appearance.
We used rigid registrations from ITK\footnote{\url{https://itk.org/Doxygen43/html/RegistrationPage.html}, as of March 2025.}, and the metrics are computed only within the phantom volume; the air areas around the phantom in the scans are ignored when computing the metrics, and the metrics are computed for liver, lung and structural parts combined.

To evaluate the stability of a quantitative image feature $x_j$ computed from various scanners, we propose to measure the similarity between features computed from CT series acquired with different imaging settings, and scanners.
Our primary goal is to focus on the influence of scanners on the structural and texture details of the reconstructed image series that may impact the feature space.
Therefore, we group the image series based on 13 scanners and compute the intra-class correlation coefficient (ICC) between the computed features from multiple ROIs of image series from various scanners.
To compute the ICC, features were averaged over reconstruction methods and repetitions.
The ICC(3,1)~\cite{shrout1979intraclass} is a commonly used measure for calculating the similarity between measurements from different sources, and is defined at feature-level as~\cite{shrout1979intraclass}
\begin{equation}
    \text{ICC(3,1)} = \frac{\textit{BMS}-\textit{EMS}}{\textit{BMS}+(i-1)\textit{EMS}},
\end{equation}
where the mean square between scanners and mean square error within scanners (i.e. ``raters'') are represented by $BMS$, and $EMS$ respectively for $i=13$ scanners in this dataset.


To evaluate the impact of scanner models and manufacturers on performance of the tissue classification task, we compute the accuracy of a multi-layer perceptron (MLP) based on three feature types: radiomics and latent representations from two DL models (details are provided in the following sections).
It should be noted that the accuracy is a suitable measure since the observations are relatively balanced across the four liver tissue classes considered.

\section*{Technical Validation}
\label{sec:validation}
In this section, we present preliminary experimental results that demonstrate the relevance of the collected dataset and provide initial baseline results for the development of harmonization methods.
In order to include classical preprocessing steps of a feature extraction pipeline, we resampled the image series volumes before feature extraction to have the same pixel spacing of $0.6836 \times 0.6836~\text{mm}$ and a slice thickness of $2~\text{mm}$.
Subsequently, we aligned the position of the phantom in all volumes of the acquired image series to maximize the structural similarity (SSIM), computed against a randomly selected volume from scanner A1 (Siemens SOMATOM Definition Edge).
These two steps were necessary to isolate the impact of the scanner on the computed results from possible spatial shifts of the phantom in the acquired images and the voxel size used for reconstruction.
To compare the impact of scanner manufacturers on quantitative image features, we consider (i) standard handcrafted radiomics features~\cite{gillies2016radiomics, rizzo2018radiomics,VFP2017} computed using the PyRadiomics\footnote{PyRadiomics: \url{https://pyradiomics.readthedocs.io/en/latest/}, as of December 2024.} library, (ii) a simple shallow convolutional neural network (CNN~\cite{lecun1998gradient}) and (iii) a vision transformer (ViT~\cite{dosovitskiy2020image}) for computing features from their latent representations.
More specifically, the latent representations are taken from the last layer of a pre-trained shallow CNN backbone~\cite{jimenez2024comparing} before the fully connected layers, and at the bottleneck of a pre-trained swin ViT-based model (SwinUNETR~\cite{Tang2022Swin}).
The shallow CNN is pre-trained on the classification of normal organ tissue in CT~\cite{jimenez2024comparing}. 
The SwinUNETR is pre-trained on a large dataset of 3D CT images using self-supervised learning.

\begin{figure}[ht!]
    \begin{center} 
    \begin{tabular}{cc}
    \subfigure[Radiomics features]{
    \includegraphics[width=0.45\textwidth]{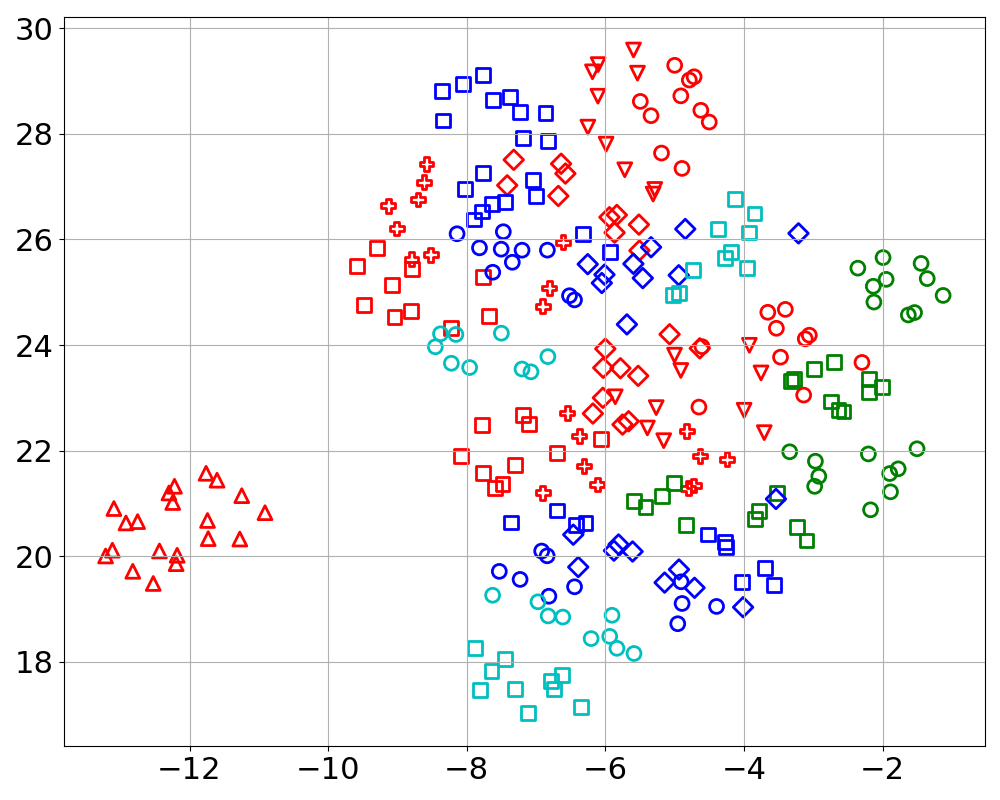}
    \label{fig:latentRadiocyst}} &
    \subfigure[CNN features]{
    \includegraphics[width=0.45\textwidth]{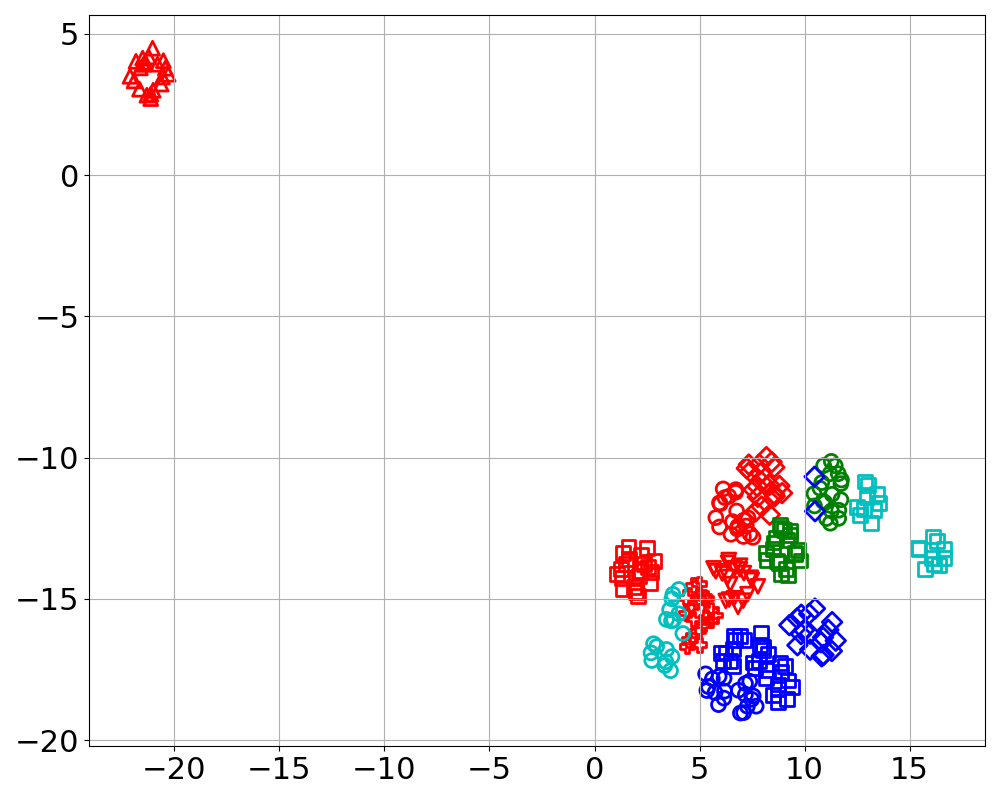}
    \label{fig:latentCNNcyst}} \\
    \subfigure[ViT features]
    {\includegraphics[width=0.45\textwidth]{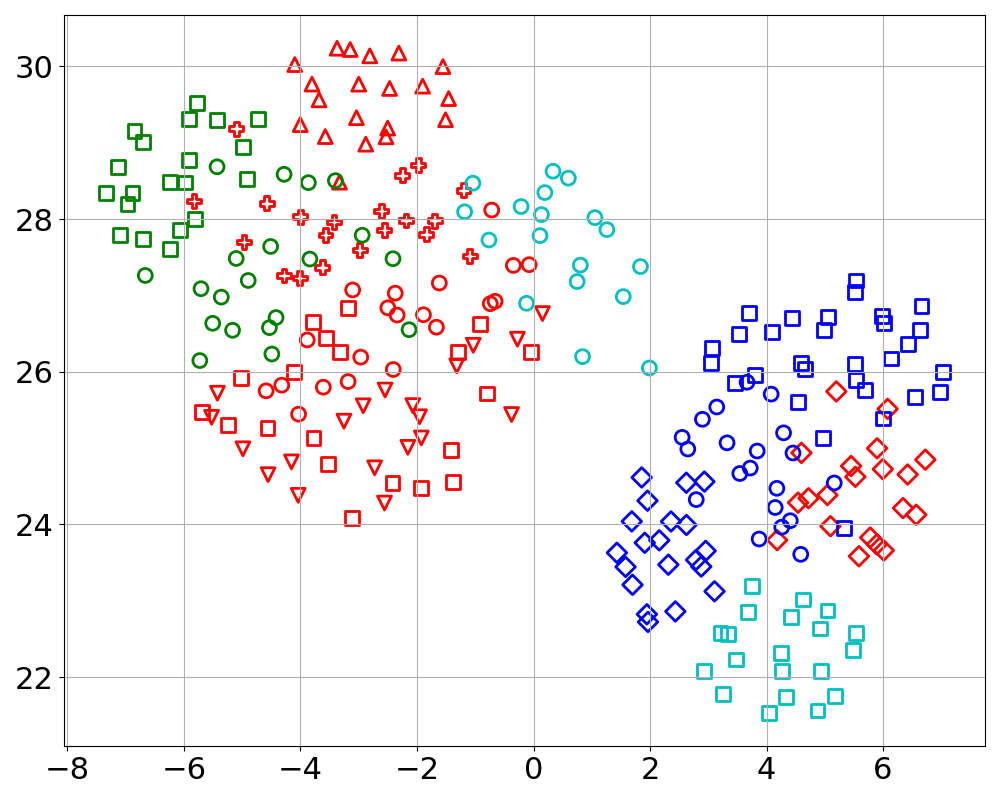}
    \label{fig:latentViTcyst}} &
    \subfigure[Legends]
    {\raisebox{0.75cm}{\includegraphics[width=0.35\textwidth]{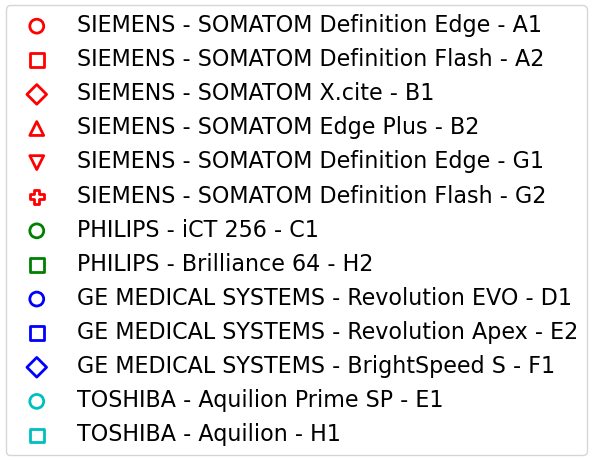}}
    \label{fig:latentcystlegend}} \\
    \end{tabular}
    \end{center}
    
    \caption{UMAP representation of radiomics features and features derived from the latent representations of the two considered computer vision models (100 neighbors and a minimum distance of 1 optimized over 100 epochs).
    This representation focuses on only one ROI—first cyst, shown in Fig.~\ref{fig:samplectcyst1}—and includes all 268 image series acquired using the harmonized protocol with a dose of 10 mGy.
    The impact of the CT scanner and manufacturer is striking with image series clustering based on the corresponding manufacturer and specific scanner model.}
    \label{fig:latentcyst}
\end{figure}

\begin{figure}[ht!]
    \begin{center} 
    \begin{tabular}{cc}
    
    {\includegraphics[height=0.028\textwidth]{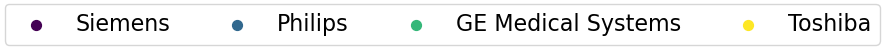}
    \label{fig:manufacturers}} &
    
    {\includegraphics[height=0.028\textwidth]{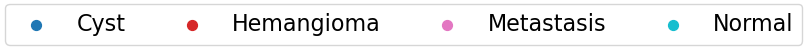}
    \label{fig:rois}}
    \\
    \subfigure[Radiomics features (color: manufacturer)]{
    \includegraphics[width=0.4\textwidth]{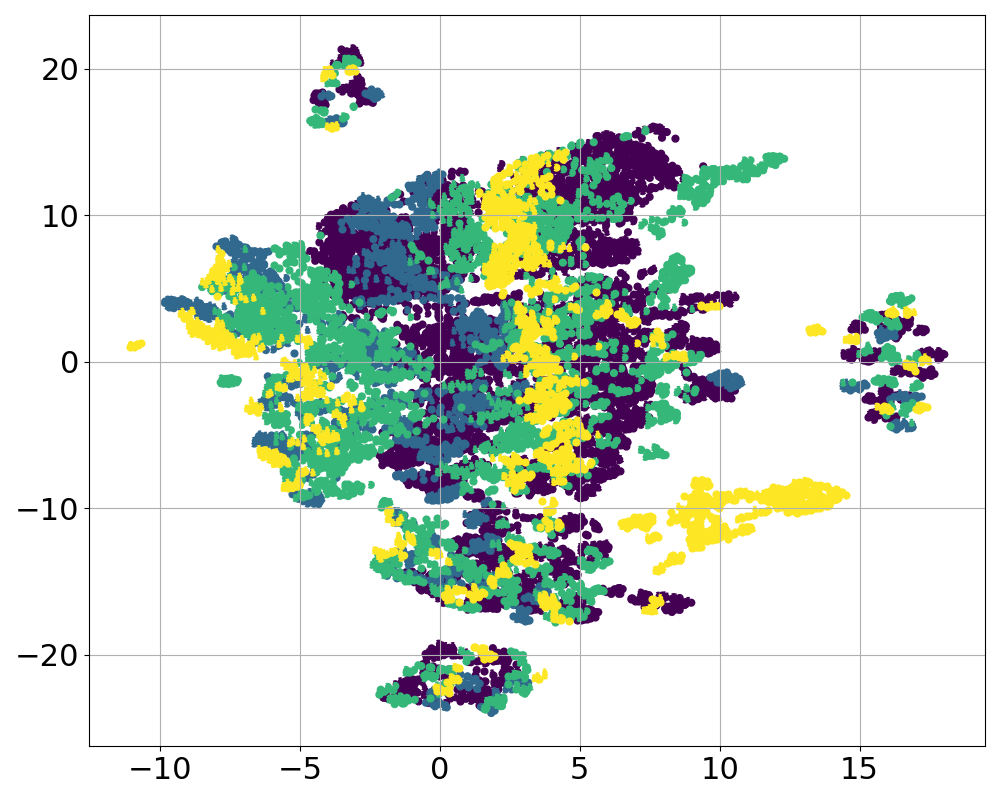}
    \label{fig:latentRadio}} &
    
    \subfigure[Radiomics features (color: liver tissue class)]{
    \includegraphics[width=0.4\textwidth]{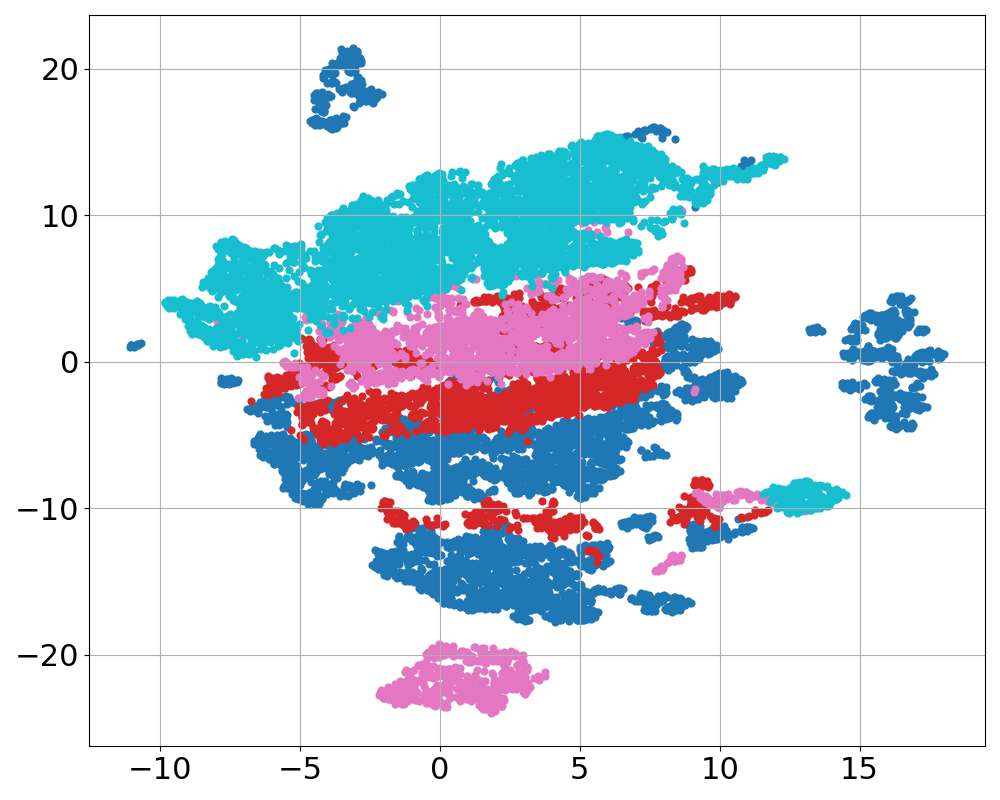}
    \label{fig:latentRadioRoi}}
    \\
    \subfigure[Shallow CNN features (color: manufacturer)]{
    \includegraphics[width=0.4\textwidth]{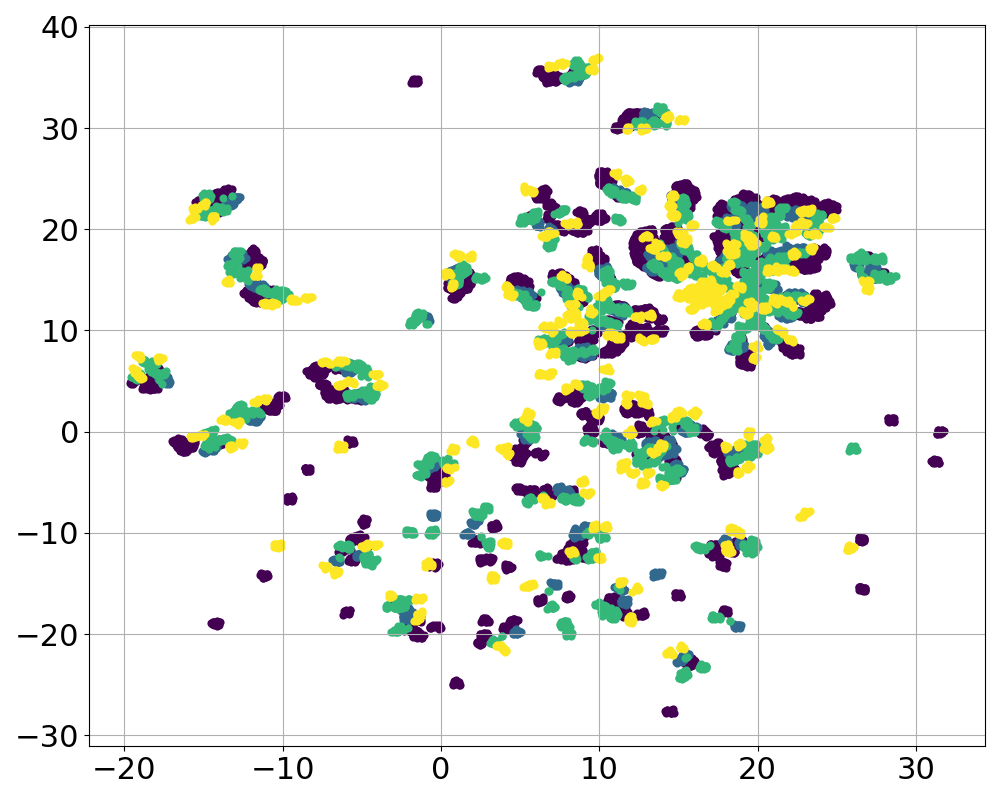}
    \label{fig:latentCNN}} &
    \subfigure[Shallow CNN features (color: liver tissue class)]{
    \includegraphics[width=0.4\textwidth]{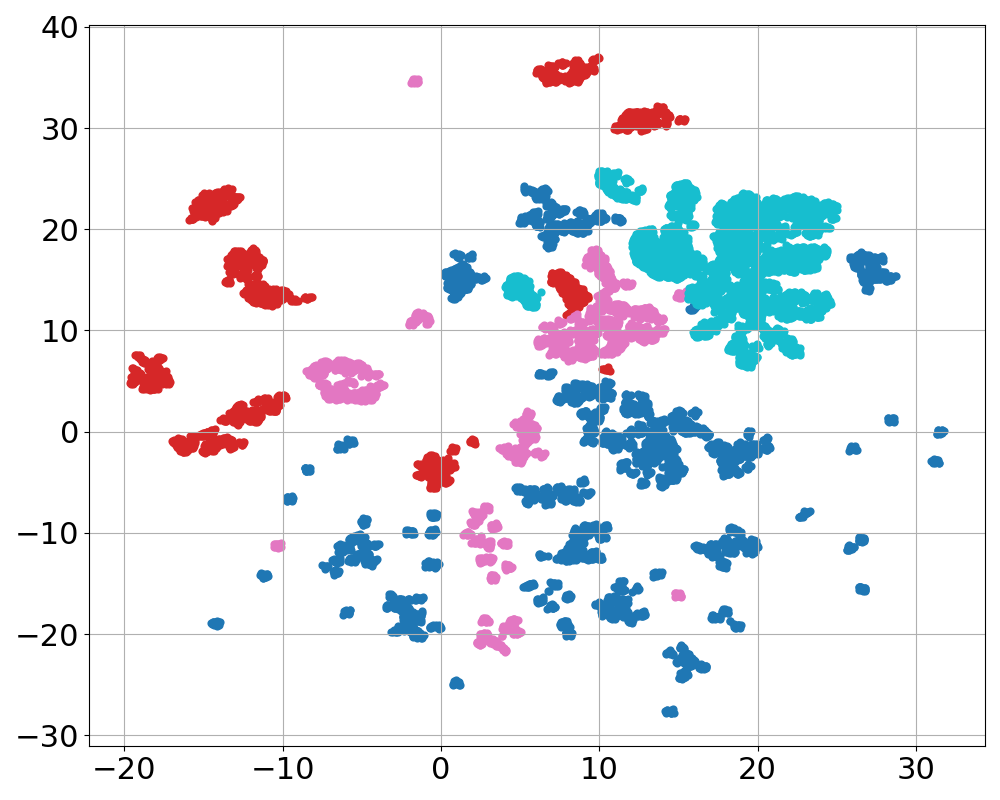}
    \label{fig:latentCNNRoi}} 
    \\
    \subfigure[SwinUNETR features (color: manufacturer)]
    {\includegraphics[width=0.4\textwidth]{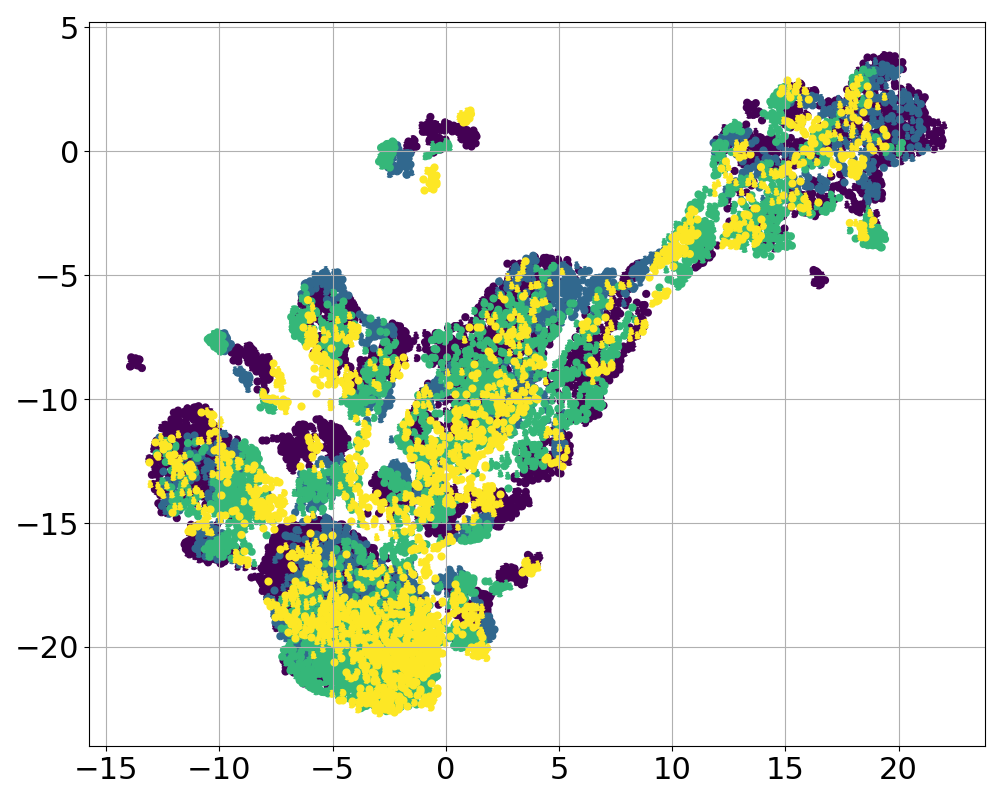}
    \label{fig:latentViT}} &
    
    \subfigure[SwinUNETR features (color: liver tissue class)]
    {\includegraphics[width=0.4\textwidth]{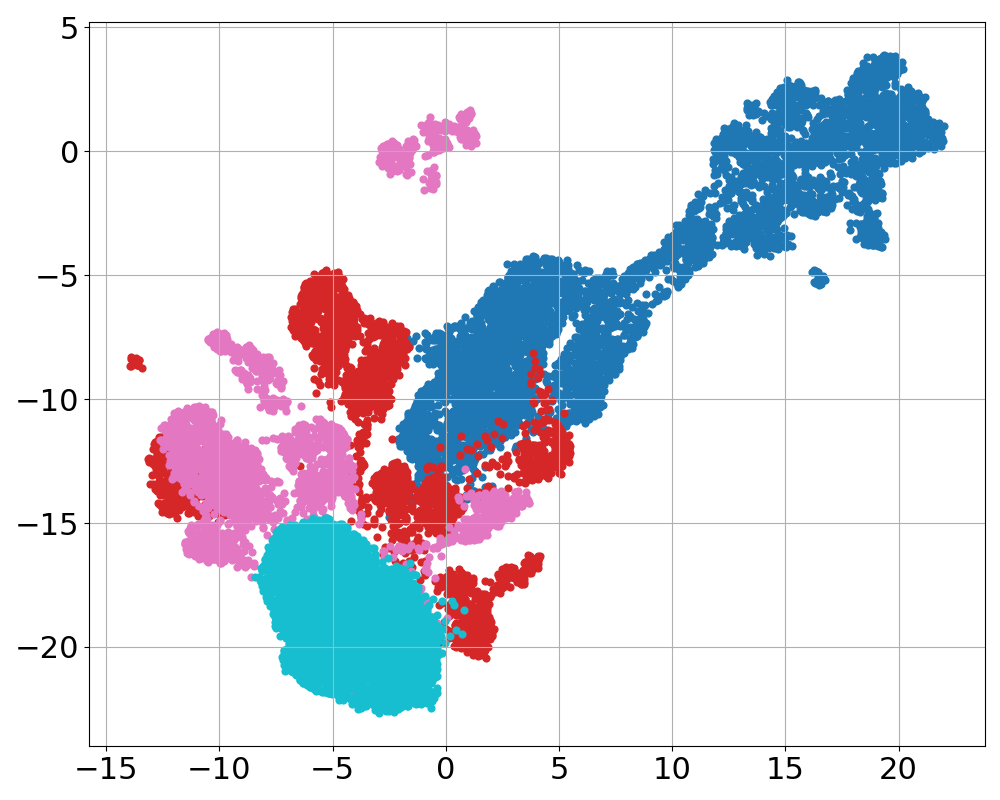}
    \label{fig:latentViTROI}} 
    
    \\
    \end{tabular}
    \end{center}
    \caption{UMAP representation of radiomics features and features derived from the latent representations of the two considered computer vision models (100 neighbors and a minimum distance of 1 optimized over 1000 epochs).
    The representation of all six ROIs from all 268 image series acquired using the harmonized protocol with a dose of 10 mGy is presented in these figures.
    The impact of the scanner manufacturer or liver tissue class on the extracted features and latent representations is illustrated with image series clustering based on the corresponding CT manufacturer (left column) or liver tissue class (right column).}
    \label{fig:latent}
\end{figure}

Figures~\ref{fig:latentcyst} and~\ref{fig:latent} present uniform manifold approximation and projection (UMAP) feature visualizations~\cite{van2012visualizing} with 100 neighbors and a minimum distance of 1 optimized over 1000 epochs highlighting the impact of scanner manufacturer on radiomics features and on the two DL latent representations.
Fig.~\ref{fig:latentcyst} focuses on features extracted from an image patch of size $32\times32\times16$ containing the first cyst tissue to depict the intra-class variations caused by scanners.
Fig.~\ref{fig:latent} shows both intra- and inter-class variations for the six tissue ROIs in the liver representing four classes: cyst (2~ROIs), hemangioma, metastasis, and normal tissue (2~ROIs), while focusing on scanner manufacturers.
In addition to the image- and feature-level visual analyses presented in Figures~\ref{fig:scanner_comparison}-\ref{fig:latent}, we present initial numerical results of the influence of scanners on image-level similarity (Tables~\ref{tab:rmse}-\ref{tab:dose}), feature-level stability and liver tissue classification performance (Table~\ref{tab:numericalresults}).
For the latter, we used a MLP with three hidden layers of sizes 100, 60, and 30, respectively, followed by a Gaussian error linear unit (GELU~\cite{hendrycks2016gaussian}) activation and trained with a dropout rate of 0.2.
The results presented in this paper are computed without applying any harmonization method providing a baseline for future research.
The details of the data split into train, validation and test sets as well as the cross-validation (CV) analyses for classification are presented in more detail in the next section.
%
Tables~\ref{tab:rmse}-\ref{tab:ssim} present image-level similarity metrics to compare the scans from different manufacturers using the harmonized protocol (with a fixed dose of 10 mGy), and various reconstruction algorithms including FBP, IR, and DL-based methods.
Table~\ref{tab:dose} details the effect of dose on image-level similarity metrics for all 13 scanners. 




The dataset introduced in this study aims to enhance research and model development for harmonization techniques in the context of CT scans acquired from different manufacturers as a primary goal, and for CT scans acquired with varying dose levels as a secondary objective.
To achieve the primary goal, the acquisition and reconstruction parameters, including the radiation dose levels were harmonized as much as possible based on a survey of commonly used protocols in Switzerland.

Fig.~\ref{fig:latentcyst} illustrates that the images are strikingly clustered based on the manufacturer in all feature types considered, potentially leading to poor inter-scanner generalization in the absence of harmonization techniques.
The intra-scanner variations presented in this figure, which are visible as a separate cluster for each single scanner, are caused by different reconstruction techniques (i.e. IR, FBP and DL), introducing an additional source of data diversity.
Fig.~\ref{fig:latent} illustrates the 2D representation of features extracted using radiomics, shallow CNN, and SwinUNETR that are colored based on the manufacturer and liver tissue type in the left and right columns, respectively. The visualizations in this figure show that the data samples are not only clustered based on tissue type but also partially according to the manufacturer of the scanners. For all computed features, the inter-class (tissue) variations seem larger than intra-class variations caused by scanner manufacturers, implying that the impact of scanners on tissue classification is limited, which is confirmed by the classification performance reported in Table~\ref{tab:numericalresults}.  

Table~\ref{tab:numericalresults} indicates a very high stability of features for radiomic features, compared to the pre-trained shallow CNN and SwinUNETR after preprocessing the image series.
In terms of tissue classification, high performances are observed, even when the images from the test scanner are not present in the training data, which is evaluated with the LOSO CV. 
As expected, the performance increases with the number of scanner seen in the training (1 versus 12 scanners with LOSO CV).
The shallow CNN features achieves best performance in tissue classification.
It is notable that scanners and manufacturers have an impact on the radiomics features and model representations as shown in Figures~\ref{fig:latentcyst} and~\ref{fig:latent}; however, this did not result in a significant degradation of the models' generalization ability to scanners that were not present in the training data.
The reported results in Table~\ref{tab:numericalresults} show that the tissue-prediction performance is higher for 10-fold CV than LOSO CV, in particular for SwinUNETR representations, highlighting the impact of including scanners similar to the test scanner samples in the training set when no harmonization method is used.
%
%
The high performance achieved by all methods for liver tissue classification also suggests that the classification task is limited in terms of complexity, where all the models performed almost perfectly.
The classification task could be easily made more difficult via the inclusion of other classes of tissue types that appear in the phantom such as organs or bone.
It is worth noting that the simple image resampling step used to unify voxel sizes acted as a basic harmonization method.

To conclude, the presented dataset allows capturing and investigating the intra- and inter-scanner differences leading to an appropriate corpus for developing and comparing data harmonization methods.
Based on all the observations, the goal of data harmonization techniques in future research is to improve the image-level metrics, ICC at the feature-level, and ultimately performances on real tasks such as tissue classification, segmentation or patient prognosis prediction through image- and feature-level data alignments.
Furthermore, image-level data harmonization techniques are expected to also increase the correlation between features in the latent space through minimizing the inconsistencies generated by various scanners.
Image-level data harmonization should focus on target tissues or structures of interest as global similarity measures do not seem to well capture inter-scanner differences as suggested by the high SSIM values in Table~\ref{tab:ssim}.
The presented dataset can be used to tackle distribution shifts directly at the image- and feature-level, whereas most recent DL methods focus on large foundation models trained on data from various sources to develop intrinsic harmonization and standardization. 

Despite the importance and recent growing attention of the community to harmonization techniques, there is an evident gap in the medical imaging niches to adapt the computer vision techniques and reduce data distribution shift.
In the context of computer vision using DL, adversarial optimization techniques~\cite{andrearczyk2019learning, andrearczyk2019neural}, contrastive learning~\cite{wang2020contrastive}, disentanglement in the latent space of representation~\cite{liu2022learning} can be used for image harmonization and unifying data distribution, especially for CT scans.
Investigating the applications of similar methods on the presented dataset in this study hints towards a very promising avenue for future research works.
The anatomical, physiological and disease variations are fixed thanks to the usage of the same phantom in all acquisitions, allowing to disentangle the latter from variations attributed to image acquisition processes.
Future work is needed to investigate how harmonization techniques developed based on this dataset can generalize to more sophisticated tasks including segmentation, diagnosis and prognosis prediction using data originating from unseen manufacturers and presenting diverse patient anatomies.
This dataset is not recommended for developing segmentation models due to the lack of diversity and simplicity of the task despite the availability of the segmentation masks. 

\section*{Usage Notes}
\label{sec:usage}
The dataset presented in this paper has a specific structure such that the CT scans are categorized based on the scanner and manufacturer used for acquisition in various institutions.
The data provided alongside this paper includes 13 folders, each labeled with a scanner code from A1 to H2 corresponding to 8 institutions, with one or two scanners used at each institution.
The encoding between letters and numbers to the scanner names and manufacturers can be deduced from the file names in the dataset, and is also provided in the first column of Table~\ref{table:scaners}.
The scans acquired from scanners produced by the same manufacturer company are more similar to each other resulting in bias in evaluation.
Thus, the training, validation, and test splits need special care and the validation scenarios have to be defined in a way to minimize the bias generated by the similarity between similar scanners from the same manufacturers.

We propose two evaluation scenarios for the classification of liver tissue types (four classes). 
We selected 10 random voxels within the 3D range of each ROI as centers, and then extracted the radiomics features and DL-based embeddings on image patches of size $32\times 32\times 16$ around these centers.
The patch centers are kept fixed across all image series.
To investigate the impact of testing on independent scanners that were not present in the training dataset, we compared it to the common scenario where scans from the test scanner are present in the training set. We define the first scenario as one in which the data from the test scanner is not present in the training set using a leave-one-scanner-out (LOSO) CV: a 13-fold CV where 13 is the number of scanners. This first scenario allows evaluating generalization abilities on unseen scanners.
For the second scenario, a standard 10-fold CV includes scans from all scanners in the training and test sets ($10\%$ of the scans kept for testing), where the models can see examples of all scanners during the training phase. 
Algorithm~1 describes the LOSO CV with a tunable parameter $N$ in the range of 1-12 to evaluate the impact of the number of scanners ($N$) used for training.
Algorithm~2 details the 10-fold CV where $n$ out of the 9 remaining training folds are used for model training.

In addition to liver tissue classification performance, we evaluate feature stability via the ICC. Table~\ref{tab:numericalresults} shows ICC values for the three feature extraction models under consideration, along with the liver tissue classification accuracy for scans acquired using the harmonized protocol and a dose of 10 mGy.
A larger ICC indicated a lower inter-scanner feature variability, which corresponds to higher feature stability for different scanners. 
Feature-level ICCs were clipped to be positive~\cite{bartko1976various} and then averaged over all features of a given extraction method to provide distinct estimation of stability for the three considered extraction methods: radiomics and two DL computer vision models (Shallow CNN and SwinUNETR).
The harmonized protocol with the fixed dose of 10 mGy is used in all classification and feature stability experiments to remove variations related to changes in dose.

The dataset is published under CC BY\footnote{\url{https://creativecommons.org/licenses/by/4.0/}, as of November 2024.} license as open source for free public usage.


~\\~\\
\textbf{Algorithm 1: Leave-one-scanner-out (LOSO) CV} 
\begin{algorithmic}[H]
\For{each scanner $s_i$ in Scanners=$\{s_1, s_2, ..., s_{13}\}$}
    \State TestData $\gets$ Pick all scans from $s_i$ for test set
    \State RemainingScanners $\gets$ Scanners - $s_i$
    \State Randomly (fix seed) pick $N$ scanners from RemainingScanners (with independent manufacturers if available)
    \State Train model with selected $N$ scanners and evaluate it on $s_i$
    \State Save performance for fold $i$ and value $N$
\EndFor
\State Compute the average performance over all 13 Scanners (for N from 1-12).
\label{algo:LOSO}
\end{algorithmic}
~\\
\textbf{Algorithm 2: 10-fold CV}
\begin{algorithmic}[H]
\For{each fold $k$ in \{1, 2, ..., 10\}}
    \State TestData $\gets$ Pick all scans of fold $k$ ($10\%$ of data) for test set
    \State RemainingData $\gets$ Data - TestData
    \State Randomly (fix seeds) pick $n$ folds out of the RemainingData for training
    \State Train model with selected training data and evaluate it on TestData
    \State Save performance for fold $k$ and data portion percentage $n$
\EndFor
\State Compute the average performance over all 10 folds (for n from 1-9).
\label{algo:10fold}
\end{algorithmic}





\section*{Code availability}
\label{sec:code}
This study is published with an open source code\footnote{Code repository: {\url{https://github.com/medgift/Harmonization-Dataset}, as of November 2024.}} to reproduce the baseline results such that researchers can start with an initial minimal implementation. 
The source code provided in conjunction with this paper includes the basic functionalities related to loading the standard train, validation, and test splits to generate the baseline results and visualizations provided in this paper.

\bibliographystyle{plain}
\bibliography{main}

\begin{thebibliography}{10}

\bibitem{AVL2014}
Hugo J W~L Aerts, Emmanuel~Rios Velazquez, Ralph T~H Leijenaar, Chintan Parmar, Patrick Grossmann, Sara Carvalho, Johan Bussink, René Monshouwer, Benjamin Haibe-Kains, Derek Rietveld, Frank Hoebers, Michelle~M Rietbergen, C~René Leemans, Andre Dekker, John Quackenbush, Robert~J Gillies, and Philippe Lambin.
\newblock Decoding tumour phenotype by noninvasive imaging using a quantitative radiomics approach.
\newblock {\em Nat Commun}, 5:4006, 6 2014.

\bibitem{amirian2021prepnet}
Mohammadreza Amirian, Javier~A Montoya-Zegarra, Jonathan Gruss, Yves~D Stebler, Ahmet~Selman Bozkir, Marco Calandri, Friedhelm Schwenker, and Thilo Stadelmann.
\newblock Prepnet: A convolutional auto-encoder to homogenize {CT} scans for cross-dataset medical image analysis.
\newblock In {\em 2021 14th International Congress on Image and Signal Processing, BioMedical Engineering and Informatics (CISP-BMEI)}, pages 1--7. IEEE, 2021.

\bibitem{andrearczyk2019learning}
Vincent Andrearczyk, Adrien Depeursinge, and Henning M{\"u}ller.
\newblock Learning cross-protocol radiomics and deep feature standardization from {CT} images of texture phantoms.
\newblock In {\em Medical Imaging 2019: Imaging Informatics for Healthcare, Research, and Applications}, volume 10954, pages 109--116. SPIE, 2019.

\bibitem{andrearczyk2019neural}
Vincent Andrearczyk, Adrien Depeursinge, and Henning M{\"u}ller.
\newblock Neural network training for cross-protocol radiomic feature standardization in computed tomography.
\newblock {\em Journal of Medical Imaging}, 6(2):024008--024008, 2019.

\bibitem{aymerich2023pilot}
Mar{\'\i}a Aymerich, Mercedes Riveira-Mart{\'\i}n, Alejandra Garc{\'\i}a-Baiz{\'a}n, Mari{\~n}a Gonz{\'a}lez-Pena, Carmen Sebasti{\`a}, Antonio L{\'o}pez-Medina, Alicia Mesa-{\'A}lvarez, Gonzalo Tard{\'a}gila de~la Fuente, Marta M{\'e}ndez-Castrill{\'o}n, Andrea Berbel-Rodr{\'\i}guez, et~al.
\newblock Pilot study for the assessment of the best radiomic features for bosniak cyst classification using phantom and radiologist inter-observer selection.
\newblock {\em Diagnostics}, 13(8):1384, 2023.

\bibitem{bach20233d}
Michael Bach, Christoph Aberle, Adrien Depeursinge, Oscar Jimenez-del Toro, Roger Schaer, Kyriakos Flouris, Ender Konukoglu, Henning M{\"u}ller, Bram Stieltjes, and Markus~M Obmann.
\newblock {3D}-printed iodine-ink {CT} phantom for radiomics feature extraction-advantages and challenges.
\newblock {\em Medical Physics}, 50(9):5682--5697, 2023.

\bibitem{bartko1976various}
John~J Bartko.
\newblock On various intraclass correlation reliability coefficients.
\newblock {\em Psychological bulletin}, 83(5):762, 1976.

\bibitem{Bashyam2020MedicalIH}
Vishnu Bashyam, Jimit Doshi, Guray Erus, Dhivya Srinivasan, Ahmed Abdulkadir, Mohamad Habes, Yong Fan, Colin~L. Masters, Paul Maruff, Chuanjun Zhuo, Henry V{\"o}lzke, Sterling~C. Johnson, Jurgen Fripp, Nikolaos Koutsouleris, Theodore~Daniel Satterthwaite, Daniel~H. Wolf, Raquel~E. Gur, Ruben~C. Gur, John~C. Morris, Marilyn~S. Albert, Hans~J{\"o}rgen Grabe, Susan~M. Resnick, R.~Nick, 17~Bryan, David~A. Wolk, Haochang Shou, Ilya~M. Nasrallah, and Christos Davatzikos.
\newblock Medical image harmonization using deep learning based canonical mapping: Toward robust and generalizable learning in imaging.
\newblock {\em ArXiv}, abs/2010.05355, 2020.

\bibitem{dosovitskiy2020image}
Alexey Dosovitskiy, Lucas Beyer, Alexander Kolesnikov, Dirk Weissenborn, Xiaohua Zhai, Thomas Unterthiner, Mostafa Dehghani, Matthias Minderer, Georg Heigold, Sylvain Gelly, Jakob Uszkoreit, and Neil Houlsby.
\newblock An image is worth 16x16 words: Transformers for image recognition at scale.
\newblock {\em International Conference on Learning Representations}, 2021.

\bibitem{Fedorov2018}
Andriy Fedorov, Michael Schwier, David Clunie, Christian Herz, Steve Pieper, Ron Kikinis, Clare Tempany, and Fiona Fennessy.
\newblock An annotated test-retest collection of prostate multiparametric {MRI}.
\newblock {\em Scientific Data 2018 5:1}, 5:1--13, 12 2018.

\bibitem{gillies2016radiomics}
Robert~J Gillies, Paul~E Kinahan, and Hedvig Hricak.
\newblock Radiomics: images are more than pictures, they are data.
\newblock {\em Radiology}, 278(2):563--577, 2016.

\bibitem{guo2022evaluation}
Lin~Lawrence Guo, Stephen~R Pfohl, Jason Fries, Alistair~EW Johnson, Jose Posada, Catherine Aftandilian, Nigam Shah, and Lillian Sung.
\newblock Evaluation of domain generalization and adaptation on improving model robustness to temporal dataset shift in clinical medicine.
\newblock {\em Scientific reports}, 12(1):2726, 2022.

\bibitem{hendrycks2021many}
Dan Hendrycks, Steven Basart, Norman Mu, Saurav Kadavath, Frank Wang, Evan Dorundo, Rahul Desai, Tyler Zhu, Samyak Parajuli, Mike Guo, et~al.
\newblock The many faces of robustness: A critical analysis of out-of-distribution generalization.
\newblock In {\em Proceedings of the IEEE/CVF international conference on computer vision}, pages 8340--8349, 2021.

\bibitem{hendrycks2016gaussian}
Dan Hendrycks and Kevin Gimpel.
\newblock Gaussian error linear units (gelus).
\newblock {\em arXiv preprint arXiv:1606.08415}, 2016.

\bibitem{hertel2023phantom}
Alexander Hertel, Hishan Tharmaseelan, Lukas~T Rotkopf, Dominik N{\"o}renberg, Philipp Riffel, Konstantin Nikolaou, Jakob Weiss, Fabian Bamberg, Stefan~O Schoenberg, Matthias~F Froelich, et~al.
\newblock Phantom-based radiomics feature test--retest stability analysis on photon-counting detector {CT}.
\newblock {\em European Radiology}, 33(7):4905--4914, 2023.

\bibitem{jahnke2017radiopaque}
Paul Jahnke, Felix~RP Limberg, Andreas Gerbl, Gracia~L Ardila~Pardo, Victor~PB Braun, Bernd Hamm, and Michael Scheel.
\newblock Radiopaque three-dimensional printing: a method to create realistic {CT} phantoms.
\newblock {\em Radiology}, 282(2):569--575, 2017.

\bibitem{jimenez2021discriminative}
Oscar Jimenez-del Toro, Christoph Aberle, Michael Bach, Roger Schaer, Markus~M Obmann, Kyriakos Flouris, Ender Konukoglu, Bram Stieltjes, Henning M{\"u}ller, and Adrien Depeursinge.
\newblock The discriminative power and stability of radiomics features with computed tomography variations: task-based analysis in an anthropomorphic {3D}-printed {CT} phantom.
\newblock {\em Investigative radiology}, 56(12):820--825, 2021.

\bibitem{jimenez2024comparing}
Oscar Jimenez-del Toro, Christoph Aberle, Roger Schaer, Michael Bach, Kyriakos Flouris, Ender Konukoglu, Bram Stieltjes, Markus~M. Obmann, André Anjos, Henning Müller, and Adrien Depeursinge.
\newblock Comparing stability and discriminatory power of hand-crafted versus deep radiomics: A 3d-printed anthropomorphic phantom study.
\newblock In {\em 2024 12th European Workshop on Visual Information Processing (EUVIP)}, pages 1--5, 2024.

\bibitem{lambin2017radiomics}
Philippe Lambin, Ralph~TH Leijenaar, Timo~M Deist, Jurgen Peerlings, Evelyn~EC De~Jong, Janita Van~Timmeren, Sebastian Sanduleanu, Ruben~THM Larue, Aniek~JG Even, Arthur Jochems, et~al.
\newblock Radiomics: the bridge between medical imaging and personalized medicine.
\newblock {\em Nature reviews Clinical oncology}, 14(12):749--762, 2017.

\bibitem{lecun1998gradient}
Yann LeCun, L{\'e}on Bottou, Yoshua Bengio, and Patrick Haffner.
\newblock Gradient-based learning applied to document recognition.
\newblock {\em Proceedings of the IEEE}, 86(11):2278--2324, 1998.

\bibitem{liu2021style}
Mengting Liu, Piyush Maiti, Sophia Thomopoulos, Alyssa Zhu, Yaqiong Chai, Hosung Kim, and Neda Jahanshad.
\newblock Style transfer using generative adversarial networks for multi-site {MRI} harmonization.
\newblock In {\em Medical Image Computing and Computer Assisted Intervention--MICCAI 2021: 24th International Conference, Strasbourg, France, September 27--October 1, 2021, Proceedings, Part III 24}, pages 313--322. Springer, 2021.

\bibitem{liu2022learning}
Xiao Liu, Pedro Sanchez, Spyridon Thermos, Alison~Q O’Neil, and Sotirios~A Tsaftaris.
\newblock Learning disentangled representations in the imaging domain.
\newblock {\em Medical Image Analysis}, 80:102516, 2022.

\bibitem{mackin2015measuring}
Dennis Mackin, Xenia Fave, Lifei Zhang, David Fried, Jinzhong Yang, Brian Taylor, Edgardo Rodriguez-Rivera, Cristina Dodge, Aaron~Kyle Jones, et~al.
\newblock Measuring computed tomography scanner variability of radiomics features.
\newblock {\em Investigative radiology}, 50(11):757--765, 2015.

\bibitem{mali2021making}
Shruti~Atul Mali, Abdalla Ibrahim, Henry~C Woodruff, Vincent Andrearczyk, Henning M{\"u}ller, Sergey Primakov, Zohaib Salahuddin, Avishek Chatterjee, and Philippe Lambin.
\newblock Making radiomics more reproducible across scanner and imaging protocol variations: a review of harmonization methods.
\newblock {\em Journal of personalized medicine}, 11(9):842, 2021.

\bibitem{marcadent2020generative}
Sandra Marcadent, Jeremy Hofmeister, Maria~Giulia Preti, Steve~P Martin, Dimitri Van De~Ville, and Xavier Montet.
\newblock Generative adversarial networks improve the reproducibility and discriminative power of radiomic features.
\newblock {\em Radiology: Artificial Intelligence}, 2(3):e190035, 2020.

\bibitem{Michallek2022}
Florian Michallek, Ulrich Genske, Stefan~Markus Niehues, Bernd Hamm, and Paul Jahnke.
\newblock Deep learning reconstruction improves radiomics feature stability and discriminative power in abdominal {CT} imaging: a phantom study.
\newblock {\em European Radiology}, 32:4587--4595, 7 2022.

\bibitem{neyshabur2017exploring}
Behnam Neyshabur, Srinadh Bhojanapalli, David McAllester, and Nati Srebro.
\newblock Exploring generalization in deep learning.
\newblock {\em Advances in neural information processing systems}, 30, 2017.

\bibitem{orlhac2022guide}
Fanny Orlhac, Jakoba~J Eertink, Anne-S{\'e}gol{\`e}ne Cottereau, Jos{\'e}e~M Zijlstra, Catherine Thieblemont, Michel Meignan, Ronald Boellaard, and Ir{\`e}ne Buvat.
\newblock A guide to combat harmonization of imaging biomarkers in multicenter studies.
\newblock {\em Journal of Nuclear Medicine}, 63(2):172--179, 2022.

\bibitem{rizzo2018radiomics}
Stefania Rizzo, Francesca Botta, Sara Raimondi, Daniela Origgi, Cristiana Fanciullo, Alessio~Giuseppe Morganti, and Massimo Bellomi.
\newblock Radiomics: the facts and the challenges of image analysis.
\newblock {\em European radiology experimental}, 2:1--8, 2018.

\bibitem{shrout1979intraclass}
Patrick~E Shrout and Joseph~L Fleiss.
\newblock Intraclass correlations: uses in assessing rater reliability.
\newblock {\em Psychological bulletin}, 86(2):420, 1979.

\bibitem{sinha2021alzheimer}
Surabhi Sinha, Sophia~I Thomopoulos, Pradeep Lam, Alexandra Muir, and Paul~M Thompson.
\newblock Alzheimer’s disease classification accuracy is improved by {MRI} harmonization based on attention-guided generative adversarial networks.
\newblock In {\em 17th international symposium on medical information processing and analysis}, volume 12088, pages 180--189. SPIE, 2021.

\bibitem{Tang2022Swin}
Yucheng Tang, Dong Yang, Wenqi Li, Holger~R. Roth, Bennett Landman, Daguang Xu, Vishwesh Nath, and Ali Hatamizadeh.
\newblock Self-supervised pre-training of swin transformers for {3D} medical image analysis.
\newblock In {\em 2022 IEEE/CVF Conference on Computer Vision and Pattern Recognition (CVPR)}, pages 20698--20708, 2022.

\bibitem{tor2020unsupervised}
Carlos Tor-Diez, Antonio~Reyes Porras, Roger~J Packer, Robert~A Avery, and Marius~George Linguraru.
\newblock Unsupervised {MRI} homogenization: application to pediatric anterior visual pathway segmentation.
\newblock In {\em Machine Learning in Medical Imaging: 11th International Workshop, MLMI 2020, Held in Conjunction with MICCAI 2020, Lima, Peru, October 4, 2020, Proceedings 11}, pages 180--188. Springer, 2020.

\bibitem{van2012visualizing}
Laurens Van~der Maaten and Geoffrey Hinton.
\newblock Visualizing non-metric similarities in multiple maps.
\newblock {\em Machine Learning}, 87(1):33--55, 2012.

\bibitem{VFP2017}
Joost J~M van Griethuysen, Andriy Fedorov, Chintan Parmar, Ahmed Hosny, Nicole Aucoin, Vivek Narayan, Regina G~H Beets-Tan, Jean-Christophe Fillion-Robin, Steve Pieper, and Hugo J W~L Aerts.
\newblock Computational radiomics system to decode the radiographic phenotype.
\newblock {\em Cancer Research}, 77:e104--e107, 2017.

\bibitem{wang2023radiomic}
Jing Wang, Boran Zhou, Xiaofeng Yang, and Tian Liu.
\newblock Radiomic feature robustness evaluations in ultrasound imaging.
\newblock In {\em Medical Imaging 2023: Ultrasonic Imaging and Tomography}, volume 12470, pages 159--164. SPIE, 2023.

\bibitem{wang2020contrastive}
Zhao Wang, Quande Liu, and Qi~Dou.
\newblock Contrastive cross-site learning with redesigned net for {COVID-19 CT} classification.
\newblock {\em IEEE Journal of Biomedical and Health Informatics}, 24(10):2806--2813, 2020.

\bibitem{wang2004image}
Zhou Wang, Alan~C Bovik, Hamid~R Sheikh, and Eero~P Simoncelli.
\newblock Image quality assessment: from error visibility to structural similarity.
\newblock {\em IEEE Transactions on Image Processing}, 13(4):600--612, 2004.

\bibitem{Zhang2023}
Jiang Zhang, Sai~Kit Lam, Xinzhi Teng, Zongrui Ma, Xinyang Han, Yuanpeng Zhang, Andy Lai~Yin Cheung, Tin~Ching Chau, Sherry Chor~Yi Ng, Francis Kar~Ho Lee, Kwok~Hung Au, Celia Wai~Yi Yip, Victor Ho~Fun Lee, Ying Han, and Jing Cai.
\newblock Radiomic feature repeatability and its impact on prognostic model generalizability: A multi-institutional study on nasopharyngeal carcinoma patients.
\newblock {\em Radiotherapy and Oncology}, 183:109578, 6 2023.

\end{thebibliography}


\section*{Acknowledgements}
This work was partly supported by the Swiss Personalized Health Network (SPHN) with the QA4IQI Quality assessment for interoperable quantitative computed tomography imaging project DMS2445 and the IMAGINE project. It was also partially supported by the Swiss National Science Foundation (SNSF, grants 325230\_197477 and 205320\_219430), the Swiss Cancer Research foundation with the project TARGET
(KFS-5549-02-2022-R) and the Lundin Family Brain Tumour Research Centre at CHUV.


\section*{Author contributions statement}

The following authors contributed to data analysis and the preparation of the tables and figures for this paper: Mohammadreza Amirian, Vincent Andrearczyk, Jean-Félix Maestrati, Maria Martin Asiain, Oscar Jimenez-del-Toro and Christoph Aberle.
The manuscript was developed and written by Mohammadreza Amirian, Vincent Andrearczyk, and Adrien Depeursinge. 
Michael Bach, Oscar Jimenez-del-Toro, Christoph Aberle,
Roger Schaer, Vincent Andrearczyk, Jean-Félix Maestrati, Kyriakos Flouris, Markus Obmann, Clarisse Dromain, Benoît Dufour, Pierre-Alexandre Alois Poletti, Hendrik von Tengg-Kobligk, Rolf H\"{u}gli, Martin Kretzschmar, Hatem Alkadhi, Ender Konukoglu, Henning M\"{u}ller, Bram Stieltjes and Adrien Depeursinge contributed to the manuscript by providing reviews and feedback.
The experiments presented in this paper for data analysis were designed and implemented by Mohammadreza Amirian, Michael Bach, Oscar Jimenez-del-Toro, Christoph Aberle, Roger Schaer, Vincent Andrearczyk, Jean-Félix Maestrati, Kyriakos Flouris, Markus Obmann, Ender Konukoglu
Henning M\"{u}ller, Bram Stieltjes and Adrien Depeursinge.
The CT data acquisition was performed by Michael Bach, Christoph Aberle, Clarisse Dromain, Beno\^{i}t Dufour, Pierre-Alexandre Alois Poletti, Hendrik von Tengg-Kobligk, Rolf H\"{u}gli, Martin Kretzschmar and Hatem Alkadhi.
The raw data was processed and prepared by Mohammadreza Amirian, Michael Bach, Oscar Jimenez-del-Toro, Christoph Aberle and Roger Schaer.


\section*{Competing interests} 
HA (Hatem Alkadhi) has received institutional grants from Bayer, Canon, Guerbet, and Siemens. HA has also received speaker honoraria from Siemens.
All authors declare no competing interests.

\section*{Figures \& Tables}

\begin{table}[ht!]
    \centering
    \rowcolors{2}{gray!25}{white}
    \resizebox{\textwidth}{!}{\begin{tabular}{C{1.1cm}|C{3cm}C{1.4cm}C{1.9cm}cC{1.5cm}C{1.5cm}C{1.7cm}C{2cm}}
\toprule
Scanner ID & Manufacturer and Model & Tube voltage [kV] &
Tube current time product [mAs]$^a$ & 
Pitch$^b$ &
Rotation time [s]$^c$ & 
Collimation [mm]~$^d$ & 
Slice thickness/ increment$^e$ & 
Reconstruction algorithm \\
\toprule
A1 & Siemens SOMATOM Definition Edge & 120 & 148 & 1.000 & 0.5 & 38.4 & 2.0~mm/ 1.0~mm & FBP$^f$~/~IR$^g$ \\ 
\midrule
A2 & Siemens SOMATOM Definition Flash & 120 & 148 & 1.000 & 0.5 & 38.4 & 2.0~mm/ 1.0~mm & FBP~/~IR \\ 
\midrule
B1 & Siemens SOMATOM X.Cite & 120 & 119 & 1.000 & 0.5 & 38.4 & 2.0~mm/ 1.0~mm & FBP~/~IR \\ 
\midrule
B2 & Siemens SOMATOM Edge Plus & 120 & 149 & 1.000 & 0.5 & 38.4 & 2.0~mm/ 1.0~mm & FBP~/~IR \\ 
\midrule
G1 & Siemens SOMATOM Definition Edge & 120 & 149 & 1.000 & 0.5 & 38.4 & 2.0~mm/ 1.0~mm & FBP~/~IR \\ 
\midrule
G2 & Siemens SOMATOM Definition Flash & 120 & 148 & 1.000 & 0.5 & 38.4 & 2.0~mm/ 1.0~mm & FBP~/~IR \\ 
\midrule
C1 & Philips Brilliance iCT 256 & 120 & 135 & 0.985 & 0.5 & 40.0 & 2.0~mm/ 1.0~mm & FBP~/~IR \\ 
\midrule
H2 & Philips Brilliance CT 64 & 120 & 153 & 1.000 & 0.5 & 40.0 & 2.0~mm/ 1.0~mm & FBP~/~IR \\ 
D1 & GE Revolution Evo$^i$ & 120 & 112.5 & 0.984 & 0.5 & 40.0 & 2.5~mm/ 1.25~mm  & FBP~/~IR/~DL$^h$ \\ 
\midrule
E2 & GE Revolution Apex & 120 & 145 & 0.984 & 0.5 & 40.0 & 2.5~mm/ 1.25~mm & FBP~/~IR/~DL \\ 
\midrule
F1 & GE BrightSpeed S & 120 & 104 & 0.938 & 0.8 & 20.0 & 2.5~mm/ 1.25~mm & FBP~/~IR \\ 
\midrule
E1 & Toshiba Aquilion Prime SP & 120 & 155 & 0.813 & 0.5 & 40.0 & 2.0~mm/ 1.0~mm & FBP~/~IR \\ 
\midrule
H1 & Toshiba Aquilion CXL & 120 & 75 & 0.828 & 0.5 & 32.0 & 2.0~mm/ 1.0~mm & FBP~/~IR \\ 
\midrule
\toprule
\end{tabular}

}
    \raggedright
    \footnotesize{$^a$ Tube current time product: adjusted such that $\text{CTDI}_{\text{vol, 32cm phantom}}$ is as close as possible to 10.0 mGy,}\\
    \footnotesize{$^b$ Pitch: as close as possible to 1.0,}\\
    \footnotesize{$^c$ Rotation time: as close as possible to 0.5 s,}\\
    \footnotesize{$^d$ Collimation: as close as possible to 40.0 mm,}\\
    \footnotesize{$^e$ The displayed field of view (FOV) was always $350$ mm,} \\
    \footnotesize{$^f$ Filtered backprojection,}\\
    \footnotesize{$^g$ Iterative reconstruction,}\\
    \footnotesize{$^h$ Deep learning based reconstruction,}\\
     \footnotesize{$^i$ GE Revolution Evo scanner's image series only contain DL-based reconstruction for doses of 1 mGy, 3 mGy, and 6 mGy, whereas only the GE Revolution Apex image series contain DL-based reconstruction for doses of 10 mGy and 14 mGy.}
    \caption{CT scanners and their scan and reconstruction parameter settings. Identical common settings for all 13 scanners included scan mode (helical, single source, single energy), a tube voltage of 120 kV, no automatic tube current modulation, a $\text{CTDI}_{\text{vol}}$, 32cm phantom of 10 mGy (with a maximum deviation of 2\%), a displayed field of view of 350 mm, and a 512x512 pixel matrix.}
    \label{table:scaners}
\end{table}
\begin{table}[ht!]
    \centering
    \rowcolors{2}{gray!25}{white}
    \resizebox{\textwidth}{!}{\begin{tabular}{ll|ccccc|ccc|cccc}
\toprule
 &  & \multicolumn{5}{c|}{Dose} & \multicolumn{3}{c|}{Reconstruction Algorithm} & Overall \\
\midrule
ID &  Manufacturer and Model & $1 mGy$ & $3 mGy$ & $6 mGy$ & $10 mGy$ & $14 mGy$ & $FBP^a$ & $IR^b$ & $DL^c$ & Series \\
\toprule
A1 & Siemens SOMATOM Definition Edge & 20 & 20 & 20 & 20 & 20 & 50 & 50 & - & 100 \\
\midrule
A2 & Siemens SOMATOM Definition Flash & 20 & 20 & 20 & 20 & 20 & 50 & 50 & - & 100 \\
\midrule
B1 & Siemens SOMATOM X.Cite & 20 & 20 & 20 & 20 & 20 & 50 & 50 & - & 100 \\
\midrule
B2 & Siemens SOMATOM Edge Plus & 20 & 20 & 20 & 20 & 20 & 50 & 50 & - & 100 \\
\midrule
G1 & Siemens SOMATOM Definition Edge & 20 & 20 & 20 & 20 & 20 & 50 & 50 & - & 100 \\
\midrule
G2 & Siemens SOMATOM Definition Flash & 20 & 20 & 20 & 20 & 20 & 50 & 50 & - & 100 \\
\midrule
C1 & Philips Brilliance iCT 256 & 20 & 20 & 20 & 20 & 20 & 50 & 50 & - & 100 \\
\midrule
H2 & Philips Brilliance CT 64 & 20 & 20 & 20 & 20 & 20 & 50 & 50 & - & 100 \\
\midrule
D1 & GE Revolution Evo & 30 & 30 & 30 & 20 & 20 & 50 & 50 & 30 & 130 \\
\midrule
E2 & GE Revolution Apex & 30 & 30 & 30 & 30 & 30 & 50 & 50 & 50 & 150 \\
\midrule
F1 & GE BrightSpeed & 20 & 20 & 20 & 20 & 20 & 50 & 50 & - & 100 \\
\midrule
E1 & Toshiba Aquilion Prime SP & 20 & 20 & 20 & 18 & 20 & 49 & 49 & - & 98 \\
\midrule
H1 & Toshiba Aquilion CXL & 20 & 20 & 20 & 20 & 20 & 50 & 50 & - & 100 \\
\midrule
\midrule
\multicolumn{2}{c|}{Sum} & 280 & 280 & 280 & 268 & 270 & 649 & 649 & 80 & 1378 \\
\toprule
\end{tabular}

}
    \raggedright
    \footnotesize{$^a$ Filtered backprojection,}\\
    \footnotesize{$^b$ Iterative reconstruction,}\\
    \footnotesize{$^c$ Deep learning based reconstruction.}
    \caption{Number of CT image series acquired from each manufacturer, categorized by acquisition dose and reconstruction algorithms.}
    \label{table:scans}
\end{table}
\begin{table}[ht!]
    \centering
    \rowcolors{2}{gray!25}{white}
    \resizebox{\textwidth}{!}{
    \begin{tabular}{>{\centering\arraybackslash}m{3cm}|*{13}{>{\centering\arraybackslash}m{2.25cm}}}
\toprule
Scanner & Siemens SOMATOM Definition Edge 
        & Siemens SOMATOM Definition Flash 
        & Siemens SOMATOM X.Cite 
        & Siemens SOMATOM Edge Plus 
        & Siemens SOMATOM Definition Edge 
        & Siemens SOMATOM Definition Flash 
        & Philips Brilliance iCT 
        & Philips Brilliance CT 64 
        & GE Revolution Evo 
        & GE Revolution Apex 
        & GE BrightSpeed 
        & Toshiba Aquilion Prime SP
        & Toshiba Aquilion CXL \\
        
\midrule 

Siemens SOMATOM Definition Edge & 7.267±2.535  \\ 
Siemens SOMATOM Definition Flash  & 44.720±0.125 & 8.694±2.374  \\
Siemens SOMATOM X.Cite  & 50.279±0.203 & 55.388±0.394 & 7.283±3.118  \\
Siemens SOMATOM Edge Plus  & 56.136±0.182 & 41.991±0.292 & 67.034±1.090 & 7.480±1.645 \\
Siemens SOMATOM Definition Edge  & 38.697±0.312 & 28.297±0.213 & 58.312±3.265 & 50.684±0.172 & 7.254±2.367  \\
Siemens SOMATOM Definition Flash  &  32.896±1.275 & 30.776±0.332 & 64.118±0.134 & 54.840±0.207 & 24.395±0.179 & 7.475±2.619 \\
Philips Brilliance iCT  & 34.613±0.446 & 42.164±0.241 & 48.433±0.276 & 63.377±0.303 & 34.587±0.297 & 33.043±0.362 & 6.726±3.088  \\
Philips Brilliance CT 64  & 45.528±0.267 & 53.225±0.240 & 28.554±0.328 & 58.792±0.169 & 35.456±0.200 & 45.329±0.297 & 44.795±0.409 & 8.264±3.115 \\
GE Revolution Evo  & 75.435±0.345 & 80.772±0.362 & 66.473±0.338 & 84.594±0.281 & 71.367±0.471 & 77.511±0.372 & 79.726±0.280 & 68.463±0.321 & 8.378±3.399 \\
GE Revolution Apex  & 58.607±0.126 & 63.002±0.208 & 46.194±0.606 & 74.481±0.366 & 60.503±0.214 & 56.616±0.214 & 51.316±0.197 & 49.406±0.496 & 69.460±0.323 & 7.864±2.717 \\
GE BrightSpeed  & 49.784±2.760 & 55.966±2.292 & 48.815±2.717 & 71.106±0.227 & 54.299±2.417 & 53.509±2.900 & 63.913±2.514 & 51.440±0.286 & 52.943±0.568 & 50.349±0.288 & 10.633±4.846 \\
Toshiba Aquilion Prime SP & 60.682±0.324 & 63.037±0.392 & 64.689±0.153 & 64.959±0.250 & 58.064±0.371 & 58.047±0.380 & 77.578±0.309 & 60.624±0.284 & 69.366±0.580 & 70.124±0.281 & 42.573±6.145 & 8.396±1.729 \\
Toshiba Aquilion CXL  & 47.006±0.368 & 55.333±0.364 & 55.063±0.296 & 79.023±0.563 & 65.299±0.526 & 50.103±0.386 & 37.302±0.429 & 59.766±0.419 & 82.833±0.385 & 70.934±3.702 & 70.738±1.035 & 68.470±0.391 & 10.173±3.888 \\

\bottomrule
\end{tabular}

}
    \caption{Image-level similarity measure: Root Mean Square Error (RMSE$\downarrow$) . The average measure is reported over multiple image series, and the standard deviation is calculated and presented next to the average values.}
    \label{tab:rmse}
\end{table}
\begin{table}[ht!]
    \centering
    \rowcolors{2}{gray!25}{white}
    \resizebox{\textwidth}{!}{
    \begin{tabular}{>{\centering\arraybackslash}m{3cm}|*{13}{>{\centering\arraybackslash}m{2.25cm}}}
\toprule
Scanner & Siemens SOMATOM Definition Edge 
        & Siemens SOMATOM Definition Flash 
        & Siemens SOMATOM X.Cite 
        & Siemens SOMATOM Edge Plus 
        & Siemens SOMATOM Definition Edge 
        & Siemens SOMATOM Definition Flash 
        & Philips Brilliance iCT 
        & Philips Brilliance CT 64 
        & GE Revolution Evo 
        & GE Revolution Apex 
        & GE BrightSpeed 
        & Toshiba Aquilion Prime SP
        & Toshiba Aquilion CXL \\
        
\midrule
Siemens SOMATOM Definition Edge & 50.449±7.264 \\ 
Siemens SOMATOM Definition Flash  & 33.011±0.024 & 47.787±3.530 \\
Siemens SOMATOM X.Cite  & 31.993±0.035 & 31.152±0.061 & 50.779±7.672 \\
Siemens SOMATOM Edge Plus  & 31.036±0.028 & 33.558±0.060 & 29.496±0.142 & 48.830±2.418 \\
Siemens SOMATOM Definition Edge  & 34.267±0.070 & 36.986±0.066 & 30.718±0.454 & 31.923±0.029 & 49.804±5.086 \\
Siemens SOMATOM Definition Flash  & 35.684±0.333 & 36.257±0.094 & 29.881±0.018 & 31.239±0.033 & 38.275±0.064 & 50.014±6.617 \\
Philips Brilliance iCT  & 35.236±0.112 & 33.522±0.050 & 32.318±0.050 & 29.982±0.041 & 35.243±0.075 & 35.640±0.095 & 51.736±8.035 \\
Philips Brilliance CT 64  & 32.855±0.051 & 31.498±0.039 & 36.908±0.100 & 30.634±0.025 & 35.027±0.049 & 32.893±0.057 & 32.996±0.079 & 49.487±7.428 \\
GE Revolution Evo  & 28.469±0.040 & 27.875±0.039 & 29.568±0.044 & 27.474±0.029 & 28.951±0.057 & 28.233±0.042 & 27.989±0.030 & 29.312±0.041 & 49.573±7.877 \\
GE Revolution Apex  & 30.662±0.019 & 30.034±0.029 & 32.730±0.114 & 28.580±0.043 & 30.385±0.031 & 30.962±0.033 & 31.816±0.033 & 32.146±0.087 & 29.186±0.040 & 49.260±5.555 \\
GE BrightSpeed  & 32.092±0.486 & 31.069±0.350 & 32.263±0.473 & 28.983±0.028 & 31.333±0.389 & 31.465±0.474 & 29.915±0.339 & 31.795±0.048 & 31.545±0.093 & 31.981±0.050 & 47.163±6.500 \\
Toshiba Aquilion Prime SP & 30.360±0.046 & 30.029±0.054 & 29.804±0.021 & 29.768±0.033 & 30.743±0.056 & 30.745±0.057 & 28.226±0.035 & 30.368±0.041 & 29.198±0.072 & 29.103±0.035 & 33.529±1.262 & 47.966±3.387 \\
Toshiba Aquilion CXL  & 32.578±0.068 & 31.161±0.057 & 31.204±0.047 & 28.066±0.062 & 29.723±0.070 & 32.024±0.067 & 34.587±0.100 & 30.492±0.061 & 27.657±0.040 & 29.017±0.488 & 29.028±0.128 & 29.311±0.050 & 47.821±8.057 \\

\bottomrule
\end{tabular}

}
    \caption{Image-level similarity measure: Peak Signal to Noise Ratio (PSNR$\uparrow$). The average measure is reported over multiple image series, and the standard deviation is calculated and presented next to the average values.}
    \label{tab:psnr}
\end{table}
\begin{table}[ht!]
    \centering
    \rowcolors{2}{gray!25}{white}
    \resizebox{\textwidth}{!}{
    \begin{tabular}{>{\centering\arraybackslash}m{3cm}|*{13}{>{\centering\arraybackslash}m{2.25cm}}}
\toprule
Scanner & Siemens SOMATOM Definition Edge 
        & Siemens SOMATOM Definition Flash 
        & Siemens SOMATOM X.Cite 
        & Siemens SOMATOM Edge Plus 
        & Siemens SOMATOM Definition Edge 
        & Siemens SOMATOM Definition Flash 
        & Philips Brilliance iCT 
        & Philips Brilliance CT 64 
        & GE Revolution Evo 
        & GE Revolution Apex 
        & GE BrightSpeed 
        & Toshiba Aquilion Prime SP
        & Toshiba Aquilion CXL \\
        
\midrule

Siemens SOMATOM Definition Edge & 0.995±0.002 \\ 
Siemens SOMATOM Definition Flash  & 0.985±0.001 & 0.994±0.002 \\
Siemens SOMATOM X.Cite  & 0.969±0.001 & 0.966±0.001 & 0.995±0.003 \\
Siemens SOMATOM Edge Plus  & 0.983±0.001 & 0.982±0.001 & 0.952±0.001 & 0.996±0.002 \\
Siemens SOMATOM Definition Edge  & 0.988±0.001 & 0.988±0.001 & 0.963±0.001 & 0.990±0.001 & 0.995±0.002 \\
Siemens SOMATOM Definition Flash  & 0.985±0.001 & 0.989±0.001 & 0.966±0.001 & 0.982±0.001 & 0.986±0.001 & 0.995±0.002 \\
Philips Brilliance iCT  & 0.978±0.001 & 0.983±0.001 & 0.970±0.000 & 0.978±0.001 & 0.979±0.000 & 0.981±0.001 & 0.996±0.002 \\
Philips Brilliance CT 64  & 0.980±0.000 & 0.979±0.001 & 0.965±0.001 & 0.977±0.001 & 0.977±0.000 & 0.979±0.000 & 0.981±0.001 & 0.994±0.003 \\
GE Revolution Evo  &  0.952±0.001 & 0.945±0.001 & 0.975±0.001 & 0.943±0.002 & 0.946±0.001 & 0.943±0.001 & 0.958±0.001 & 0.955±0.001 & 0.995±0.003 \\
GE Revolution Apex  & 0.971±0.001 & 0.971±0.001 & 0.974±0.001 & 0.968±0.001 & 0.971±0.000 & 0.968±0.001 & 0.977±0.001 & 0.974±0.001 & 0.977±0.001 & 0.995±0.002 \\
GE BrightSpeed  & 0.953±0.001 & 0.948±0.001 & 0.970±0.001 & 0.947±0.001 & 0.949±0.001 & 0.948±0.001 & 0.962±0.001 & 0.960±0.001 & 0.983±0.001 & 0.973±0.001 & 0.993±0.004  \\
Toshiba Aquilion Prime SP & 0.983±0.001 & 0.982±0.001 & 0.975±0.001 & 0.980±0.001 & 0.981±0.001 & 0.983±0.001 & 0.981±0.001 & 0.979±0.001 & 0.966±0.002 & 0.978±0.001 & 0.964±0.001 & 0.994±0.002 \\
Toshiba Aquilion CXL  & 0.941±0.001 & 0.933±0.001 & 0.968±0.001 & 0.926±0.001 & 0.933±0.001 & 0.933±0.001 & 0.950±0.001 & 0.938±0.001 & 0.971±0.001 & 0.960±0.001 & 0.979±0.001 & 0.948±0.001 & 0.992±0.004 \\

\bottomrule
\end{tabular}

}
    \caption{Image-level similarity measure: Structural Similarity (SSIM$\uparrow$).  The average measure is reported over multiple image series, and the standard deviation is calculated and presented next to the average values.}
    \label{tab:ssim}
\end{table}
\begin{table}[ht!]
    \centering
    \rowcolors{2}{gray!25}{white}
    \resizebox{!}{0.63\textwidth}{

\begin{tabular}{lc|cccc}
    \toprule
    \textbf{Scanner} & \textbf{Dose} & \textbf{RMSE$\downarrow$} & \textbf{PSNR$\uparrow$} & \textbf{SSIM$\uparrow$} \\
    \midrule
    Siemens SOMATOM Definition Edge & 1 mGy & 39.140±4.801 & 34.239±1.131 & 0.896±0.021 \\ 
     & 3 mGy & 30.259±4.241 & 36.487±1.199 & 0.932±0.009 \\ 
     & 6 mGy & 29.703±2.790 & 36.604±0.826 & 0.951±0.008  \\ 
     & 10 mGy & 21.038±2.821 & 39.648±1.270 & 0.966±0.007 \\ 
    \midrule
    Siemens SOMATOM Definition Edge & Average & 29.500±6.278 & 36.817±1.828 & 0.938±0.022  \\ 
    \bottomrule  
    Siemens SOMATOM Definition Flash & 1 mGy & 66.659±28.267 & 30.315±3.628 & 0.789±0.091 \\ 
     & 3 mGy & 43.238±3.015 & 33.325±0.609 & 0.906±0.014  \\ 
     & 6 mGy & 31.340±2.983 & 36.140±0.862 & 0.945±0.008 \\ 
     & 10 mGy & 30.579±4.898 & 36.415±1.311 & 0.958±0.008 \\ 
    \midrule
    Siemens SOMATOM Definition Flash & Average & 38.212±8.482 & 34.597±1.970 & 0.916±0.041 \\ 
    \bottomrule   
    Siemens SOMATOM X.Cite & 1 mGy & 47.550±3.370 & 32.499±0.614 & 0.867±0.017  \\ 
     & 3 mGy & 28.964±2.706 & 36.821±0.801 & 0.942±0.009  \\ 
     & 6 mGy & 36.809±4.137 & 34.763±1.067 & 0.954±0.005  \\ 
     & 10 mGy & 29.551±8.626 & 37.024±2.755 & 0.963±0.006  \\ 
    \midrule
    Siemens SOMATOM X.Cite & Average & 36.552±6.898 & 34.917±1.638 & 0.929±0.041   \\  
    \bottomrule
    Siemens SOMATOM Edge Plus & 1 mGy & 41.632±6.065 & 33.737±1.392 & 0.911±0.019   \\ 
     & 3 mGy & 39.878±5.331 & 34.091±1.243 & 0.938±0.009 \\ 
     & 6 mGy & 31.500±9.696 & 36.534±2.990 & 0.961±0.011 \\ 
     & 10 mGy & 33.991±10.078 & 35.977±3.532 & 0.968±0.008  \\ 
    \midrule
    Siemens SOMATOM Edge Plus & Average &  40.021±2.486 & 33.992±0.540 & 0.946±0.018   \\ 
    \bottomrule    
    Siemens SOMATOM Definition Edge & 1 mGy & 44.144±6.259 & 33.214±1.268 & 0.894±0.021  \\ 
     & 3 mGy & 36.678±3.140 & 34.764±0.737 & 0.931±0.010 \\ 
     & 6 mGy & 32.652±7.242 & 35.979±2.079 & 0.953±0.010 \\ 
     & 10 mGy & 31.732±5.923 & 36.144±1.638 & 0.963±0.005 \\ 
    \midrule
    Siemens SOMATOM Definition Edge & Average & 36.642±6.540 & 34.866±1.438 & 0.939±0.025 \\ 
    \bottomrule    
    Siemens SOMATOM Definition Flash & 1 mGy & 40.303±4.160 & 33.962±0.919 & 0.885±0.019 \\ 
     & 3 mGy & 34.583±3.646 & 35.297±0.995 & 0.926±0.011 \\ 
     & 6 mGy & 32.534±1.223 & 35.780±0.328 & 0.946±0.008 \\ 
     & 10 mGy & 26.123±5.469 & 37.884±1.915 & 0.961±0.005 \\ 
    \midrule
    Siemens SOMATOM Definition Flash & Average & 35.227±5.541 & 35.189±1.353 & 0.925±0.034 \\ 
    \bottomrule    
    Philips Brilliance iCT 256 & 1 mGy & 73.283±38.214 & 29.898±4.469 & 0.738±0.153  \\ 
     & 3 mGy & 40.749±4.918 & 33.882±1.055 & 0.879±0.030  \\ 
     & 6 mGy & 30.144±2.205 & 36.460±0.637 & 0.932±0.009 \\ 
     & 10 mGy & 23.999±2.341 & 38.461±0.892 & 0.948±0.008 \\ 
    \midrule
    Philips Brilliance iCT 256 & Average & 35.201±7.651 & 35.304±1.952 & 0.899±0.041 \\ 
    \bottomrule    
    Philips Brilliance CT 64 & 1 mGy & 130.541±0.801 & 23.706±0.053 & 0.476±0.002 \\ 
     & 3 mGy & 38.385±5.664 & 34.432±1.279 & 0.879±0.029 \\ 
     & 6 mGy & 31.945±1.670 & 35.944±0.444 & 0.917±0.009 \\ 
     & 10 mGy & 28.872±2.718 & 36.850±0.828 & 0.932±0.008  \\ 
    \midrule
    Philips Brilliance CT 64 & Average & 56.781±42.621 & 32.852±5.328 & 0.807±0.192 \\ 
    \bottomrule    
    GE Revolution Evo & 1 mGy & 36.847±1.530 & 34.700±0.353 & 0.907±0.011 \\ 
     & 3 mGy & 37.370±11.357 & 34.972±2.638 & 0.900±0.056 \\ 
     & 6 mGy & 31.020±6.107 & 36.335±1.541 & 0.935±0.025 \\ 
     & 10 mGy & 28.527±2.101 & 36.941±0.675 & 0.944±0.011 \\ 
    \midrule
    GE Revolution Evo & Average & 31.380±3.075 & 36.127±0.824 & 0.932±0.014 \\ 
    \bottomrule    
    GE Revolution Apex & 1 mGy & 68.446±2.879 & 29.321±0.360 & 0.906±0.027 \\ 
     & 3 mGy & 63.843±1.478 & 29.921±0.201 & 0.940±0.012  \\ 
     & 6 mGy & 47.533±3.015 & 32.498±0.556 & 0.964±0.011 \\ 
     & 10 mGy & 45.680±1.592 & 32.831±0.299 & 0.960±0.010 \\ 
    \midrule
    GE Revolution Apex & Average & 56.116±9.513 & 31.166±1.489 & 0.947±0.015 \\ 
    \bottomrule    
    GE BrightSpeed & 1 mGy & 138.336±6.911 & 23.213±0.437 & 0.605±0.060 \\ 
     & 3 mGy & 109.802±30.469 & 25.635±2.894 & 0.786±0.088 \\ 
     & 6 mGy & 58.964±15.550 & 30.856±1.966 & 0.913±0.014 \\ 
     & 10 mGy & 56.299±16.765 & 31.325±2.217 & 0.933±0.007 \\ 
    \midrule
    GE BrightSpeed & Average & 92.996±42.814 & 27.678±4.303 & 0.782±0.153 \\ 
    \bottomrule    
    Toshiba Aquilion Prime SP & 1 mGy & 89.800±44.376 & 28.444±5.481 & 0.676±0.215 \\ 
     & 3 mGy & 45.444±11.811 & 33.211±2.519 & 0.878±0.063 \\ 
     & 6 mGy & 31.101±7.788 & 36.457±2.287 & 0.937±0.026 \\ 
     & 10 mGy & 33.136±4.964 & 35.718±1.356 & 0.947±0.015 \\ 
    \midrule
    Toshiba Aquilion Prime SP & Average & 61.785±37.589 & 31.565±4.636 & 0.805±0.180  \\ 
    \bottomrule
    Toshiba Aquilion CXL & 1 mGy & 102.640±61.002 & 27.362±5.173 & 0.648±0.289 \\ 
     & 3 mGy & 75.196±50.152 & 30.537±5.929 & 0.761±0.203  \\ 
     & 6 mGy & 34.122±10.842 & 35.802±2.782 & 0.911±0.050 \\ 
     & 10 mGy & 34.176±8.515 & 35.687±2.581 & 0.912±0.031 \\ 
    \midrule
    Toshiba Aquilion CXL & Average & 41.799±9.338 & 33.790±1.777 & 0.890±0.025 \\ 
    \bottomrule

\end{tabular}

}
    \caption{Image-level similarity measures at different dose levels, with the highest dose of 14 mGy considered as the reference volume.}
    \label{tab:dose}
\end{table}

\begin{table}[ht!]
    \centering
    
    \begin{tabular}{l|c|cc|c}
     & Feature stability &  \multicolumn{3}{c}{Liver tissue classification} \\
    \midrule
    \multirow{2}{10em}{Features}   &
    \multirow{2}{*}{Mean ICC}  &  \multicolumn{3}{c}{Mean CV accuracy}    \\
    &  & \multicolumn{2}{c}{LOSO} & 10-fold \\
    Training scanners & & 1 scanner & 12 scanners & 13 scanners \\
    \midrule
    Radiomics~\cite{VFP2017}                      & 0.990±0.121 & 0.920±0.027 & 0.986±0.024 & 0.997±0.001 \\
    Shallow CNN~\cite{jimenez2024comparing}       & 0.965±0.103 & 0.949±0.035 & 0.998±0.005 & 1.000±0.000 \\
    SwinUNETR~\cite{Tang2022Swin}                 & 0.917±0.182 & 0.819±0.082 & 0.985±0.026 & 0.998±0.002 \\ 
    \toprule
    \end{tabular}
    \caption{Average feature stability and liver tissue classification performance based on scans acquired using the harmonized protocol (dose of 10 mGy). For tissue classification, performances using LOSO and 10-fold CVs are reported, where LOSO strictly evaluates the performance on unseen scanners.
    Standard deviations ($\pm$) are provided for all averaged measures.}
    \label{tab:numericalresults}
\end{table}

\end{document}